\documentclass[11pt]{article}

\usepackage[preprint]{acl}

\usepackage{times}
\usepackage{latexsym}

\usepackage[T1]{fontenc}
\usepackage[utf8]{inputenc}

\usepackage{microtype}

\usepackage{inconsolata}

\usepackage{graphicx}

\usepackage[table]{xcolor}  % 放在导言区
\usepackage{listings}
\usepackage{titletoc}
\usepackage{graphicx}
\usepackage{booktabs}
\usepackage{amsmath}
\usepackage[noabbrev,capitalise]{cleveref}

\usepackage{amssymb}
\usepackage{multicol}
\usepackage{multirow}
\usepackage{xspace}
\usepackage{makecell}
\usepackage{bm}
\usepackage{subfigure}
\usepackage{tikz}
\usepackage{tcolorbox}
\usepackage{enumitem}
\usepackage{url}
\usepackage{hyperref}

\newcommand{\eg}{\emph{e.g.,}\xspace}

\newcommand{\ignore}[1]{}
\newcommand{\paratitle}[1]{\vspace{1.5ex}\noindent\textbf{#1}}

\title{DeepInterestGR: Mining Deep Multi-Interest Using Multi-Modal LLMs for Generative Recommendation}

\author{Yangchen Zeng \\
	\texttt{zengyangchen@foxmail.com} \\\And
	guangyuan dong \\
	\texttt{} \\\And
	Zhenyu Yu \\
	\texttt{} \\\And
	Zhiyuan Hu \\
	\texttt{} \\\And
	Jinze Wang\thanks{Corresponding author} \\
	\texttt{jinzewang@swin.edu.au} \\\And
	Rongfeng Guo \\
	\texttt{} \\
}

\begin{document}
	\maketitle
	\begin{abstract}
		We introduce DeepInterestGR, a novel framework that integrates deep interest mining into the generative recommendation pipeline.
		This addresses the ``Shallow Interest'' problem---existing generative methods rely on surface-level textual features and fail to capture latent user motivations, limiting personalization depth and recommendation interpretability.
		Our approach leverages Multi-LLM Interest Mining (MLIM) via structured reasoning prompting, Reward-Labeled Deep Interest (RLDI) for quality control, and Interest-Enhanced Item Discretization (IEID) via RQ-VAE, combined with a two-stage SFT-GRPO training pipeline guided by an Interest-Aware Reward.
		We validate DeepInterestGR on three Amazon Review benchmarks (Beauty, Sports, Instruments), comparing against 14 state-of-the-art baselines including SASRec, BERT4Rec, TIGER, LC-Rec, and S-DPO.
		Our method achieves 12.4\%--15.4\% relative improvements on HR@10 and 15.5\%--17.9\% on NDCG@10, with cross-domain generalization gains of +24.8\%. To our knowledge, this is the first work to demonstrate such significant gains by incorporating deep semantic interests into generative recommendation.
		\textbf{Anonymous Link:} \url{https://anonymous.4open.science/r/generativeRec_EMNLP26}
	\end{abstract}
	
	\begin{figure}[t]
		\centering
		\includegraphics[width=0.55\textwidth]{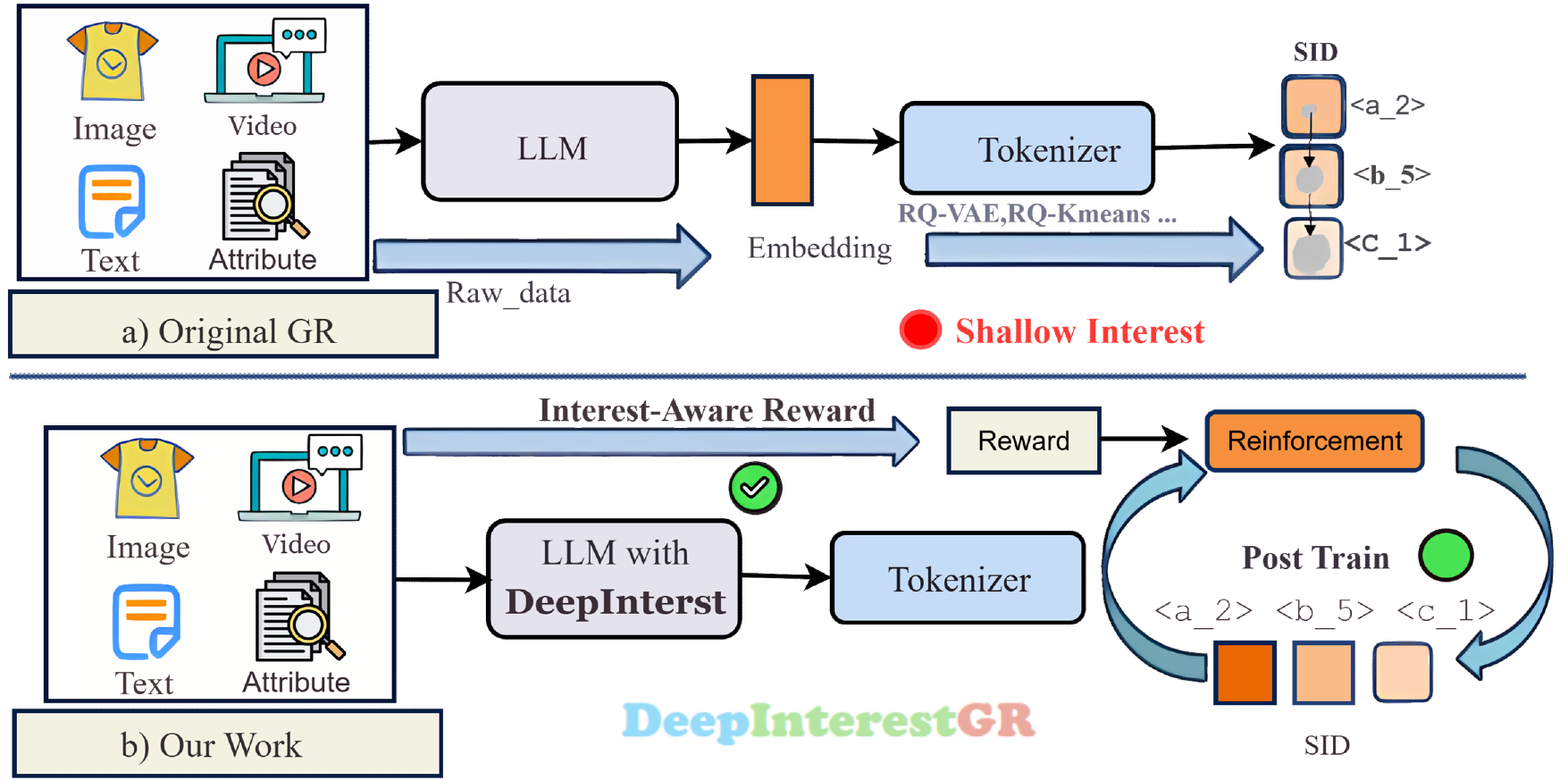}
		\caption{Comparison of traditional generative recommendation (top) and DeepInterestGR (bottom). Traditional methods suffer from ``Shallow Interest'' by only using surface-level features, while DeepInterestGR incorporates deep interest mining and Interest-Aware Reward for reinforcement learning optimization.}
		\label{fig:comparison}
	\end{figure}
	
	\vspace{-4mm}
	\section{Introduction}
	
	The scaling behavior of recommendation models has recently attracted significant attention~\cite{zhai2024hstu,li2024grsurvey,li2024grsurvey2,kong2025minionerec}, driven largely by the success of large language models (LLMs) in demonstrating predictable performance gains with increased model size~\cite{zhao2023llm_survey,devlin2019bert}. However, traditional recommendation models have struggled to exhibit similar scaling laws~\cite{kang2018sasrec,hidasi2016gru4rec,tang2018caser,sun2019bert4rec}. These systems typically allocate the majority of parameters to large embedding tables for storing user and item representations, while using only inner-product or shallow scoring networks for final predictions. Such an embedding-heavy design leads to performance plateaus: even as embedding dimensions or table sizes increase, improvements diminish quickly beyond moderate scales~\cite{hou2022unisrec,li2017narm,wu2019srgnn}.
	
	Generative recommendation offers a fundamental paradigm shift. By compressing items into sequences of discrete Semantic IDs (SIDs) through quantization techniques~\cite{jegou2011pq,ge2014opq}, it uses compact vocabularies and redirects the bulk of parameters toward deep autoregressive Transformers that generate SID sequences~\cite{rajput2023tiger,zheng2024lcrec,geng2022p5,petrov2023gptrec}. This design enables scaling behaviors more characteristic of language models, where increased depth and capacity translate to consistent performance gains~\cite{zhai2024hstu,hou2025generative}. Recent works such as TIGER~\cite{rajput2023tiger}, LC-Rec~\cite{zheng2024lcrec}, and industrial systems have demonstrated the promise of this generative paradigm.
	
	Despite these advances, existing generative recommendation methods suffer from a critical limitation: they primarily rely on \textbf{shallow behavioral signals}. Current approaches encode items solely through surface-level textual features such as titles and descriptions~\cite{rajput2023tiger,zheng2024lcrec,hou2023vqrec}, failing to capture the latent, semantically rich interests underlying user interactions. We term this the ``\textbf{Shallow Interest}'' problem, which is illustrated in \Cref{fig:comparison}. 
	
	To address these limitations, we propose \textbf{DeepInterestGR}, a novel framework that integrates deep interest mining into the generative recommendation pipeline. Our key insight is that frontier LLMs possess remarkable world knowledge and reasoning capabilities~\cite{zhao2023llm_survey,li2023recformer}, which can be leveraged to extract interpretable, semantically rich interest representations from user behaviors.
	
	DeepInterestGR introduces three key innovations:
	\begin{itemize}
		\item \textbf{Multi-LLM Interest Mining (MLIM).} We systematically leverage multiple frontier LLMs via API (\eg GPT, Gemini, Kimi, Grok) along with their multi-modal variants to extract deep textual and visual interest representations for users and items through structured reasoning prompting. Through comparative experiments, we identify the optimal LLM configuration for high-quality interest mining.
		\item \textbf{Reward-Labeled Deep Interest (RLDI).} To ensure interest quality for downstream reinforcement learning, we employ a lightweight binary classifier (based on Qwen2.5-7B-Instruct) to assign reward labels (positive/negative) to mined interests. These labels serve as supervision signals during the RL phase.
		\item \textbf{Interest-Enhanced Item Discretization (IEID).} The curated deep interests are encoded into semantic embeddings via Qwen-Embedding and quantized into SID tokens via RQ-VAE, enriching item representations with interpretable interest semantics.
	\end{itemize}
	
	For training, we adopt a two-stage pipeline following the successful practice of generative recommendation~\cite{rajput2023tiger,zheng2024lcrec}: supervised fine-tuning (SFT) aligns the generative model with both deep interest signals and collaborative filtering patterns, followed by reinforcement learning. Crucially, we introduce an \textbf{Interest-Aware Reward} derived from RLDI labels---recommendations consistent with positive interests receive higher rewards, guiding the policy toward more personalized and semantically coherent item sequences.
	
	Extensive experiments on large-scale real-world benchmarks from the Amazon Review Dataset~\cite{mcauley2015amazon} demonstrate that DeepInterestGR achieves significant improvements over state-of-the-art baselines including SASRec~\cite{kang2018sasrec}, BERT4Rec~\cite{sun2019bert4rec}, TIGER~\cite{rajput2023tiger}, LC-Rec~\cite{zheng2024lcrec}, and VQ-Rec~\cite{hou2023vqrec} on HR@K and NDCG@K metrics, validating the effectiveness of unifying deep interest mining with generative recommendation.
	
	Our contributions include: (1) identifying the ``Shallow Interest'' problem and proposing DeepInterestGR to integrate deep interest mining into generative recommendation; (2) introducing MLIM, RLDI, and IEID as core components; (3) designing an Interest-Aware Reward for RL optimization; and (4) validating state-of-the-art performance across benchmarks.

	Along this line, our prior work spans generative recommendation, multimodal reasoning, and robust visual perception~\cite{zeng2026deepinterestgr,zeng2026trialigngr,zeng2026deep,zeng2026learning,wang2026agent4poi,zeng2025hmpe,yu2026cast,hong2026uv,yu2026ads,hong2026anomaly,yu2026can,wang2026meta}.
	
	\section{Related Work}
	
	\noindent\textbf{Sequential Recommendation.} Traditional methods include Markov chains~\cite{rendle2010fpmc}, RNNs~\cite{hidasi2016gru4rec}, and attention-based models like SASRec~\cite{kang2018sasrec} and BERT4Rec~\cite{sun2019bert4rec}. However, embedding-heavy designs limit scalability~\cite{hou2022unisrec}.
	
	\noindent\textbf{Generative Recommendation.} Recent works reformulate recommendation as sequence generation using SID quantization~\cite{li2024grsurvey,li2024grsurvey2,hou2025generative}. TIGER~\cite{rajput2023tiger} pioneered RQ-VAE-based SID generation, followed by LC-Rec~\cite{zheng2024lcrec}, VQ-Rec~\cite{hou2023vqrec}, EAGER~\cite{wang2024eager}, LETTER~\cite{wang2024letter}, CoST~\cite{zhu2024cost}, and HSTU~\cite{zhai2024hstu}. However, these methods rely on shallow textual features, missing latent user interests.
	
	\noindent\textbf{LLM for Recommendation.} LLMs have been applied via prompt-based learning~\cite{geng2022p5}, sequential modeling~\cite{li2023recformer}, and textual ID learning~\cite{tan2024idgenrec,petrov2023gptrec,hou2022unisrec}, with RL optimization using behavioral signals~\cite{lin2024efficient}.
	
	\begin{figure*}[t]
		\centering
		\includegraphics[width=0.95\textwidth]{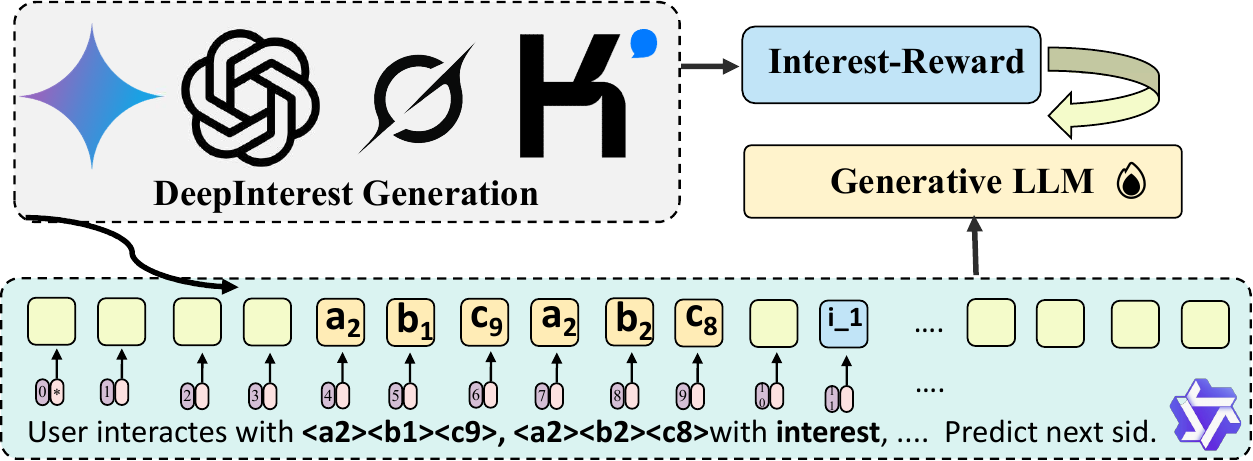}
		\caption{DeepInterestGR workflow overview. User interactions are encoded as SID tokens, which are processed through Multi-LLM Interest Mining (GPT, Gemini, Kimi, Grok) to generate deep interest representations. These interests compute the Interest-Aware Reward that guides the Generative LLM for autoregressive next-SID prediction, forming a closed-loop optimization process. See Appendix \Cref{fig:framework} for detailed architecture including IEID encoding and two-stage training.}
		\label{fig:pipeline}
	\end{figure*}
	
	\section{Method}\label{sec:method}

	In this section, we present the DeepInterestGR framework in detail. We first formalize the problem setting and introduce the notation used throughout this paper (\Cref{sec:problem}). We then describe the generative recommendation paradigm that serves as the foundation of our approach (\Cref{sec:gr_framework}). Subsequently, we present our core contributions: Multi-LLM Interest Mining (MLIM) for extracting deep interests (\Cref{sec:mlim}), Interest-Enhanced Item Discretization (IEID) for enriching item representations (\Cref{sec:ieid}), and our two-stage training pipeline with Interest-Aware Reward (\Cref{sec:training}). An overview of the complete workflow is illustrated in \Cref{fig:pipeline}.
	
	\subsection{Problem Formulation}\label{sec:problem}
	
	We consider the sequential recommendation task, where the goal is to predict the next item a user will interact with based on their historical interaction sequence. Let $\mathcal{U}$ denote the set of users and $\mathcal{I}$ denote the set of items, with $|\mathcal{U}| = N$ and $|\mathcal{I}| = M$ representing the number of users and items, respectively.
	
	\paratitle{User Interaction Sequence.}
	For each user $u \in \mathcal{U}$, we observe a chronologically ordered interaction sequence $\mathcal{S}_u = [i_1, i_2, \ldots, i_T]$, where $i_t \in \mathcal{I}$ represents the item that user $u$ interacted with at time step $t$, and $T$ denotes the sequence length. The sequential recommendation task aims to predict the next item $i_{T+1}$ that user $u$ is likely to interact with, given the historical sequence $\mathcal{S}_u$.
	
	\paratitle{Item Representation.}
	Each item $i \in \mathcal{I}$ is associated with rich metadata, including textual attributes (\eg title $\mathbf{t}_i$, description $\mathbf{d}_i$, category $\mathbf{c}_i$) and visual content (\eg product image $\mathbf{v}_i$). Existing generative recommendation methods typically encode items using only surface-level textual features:
	\begin{equation}
		\mathbf{e}_i^{\text{shallow}} = f_{\text{enc}}(\mathbf{t}_i, \mathbf{d}_i),
	\end{equation}
	where $f_{\text{enc}}(\cdot)$ denotes a text encoder (\eg sentence transformer). This shallow encoding captures explicit attributes but fails to reveal the latent interests underlying user-item interactions.
	
	\paratitle{Deep Interest.}
	We introduce the concept of \textit{deep interest} to capture the semantically rich, latent motivations behind user behaviors. For each item $i$, we define the deep interest representation as:
	\begin{equation}
		\mathbf{z}_i = \text{MLIM}(\mathbf{t}_i, \mathbf{d}_i, \mathbf{v}_i),
	\end{equation}
	where $\text{MLIM}(\cdot)$ denotes our Multi-LLM Interest Mining module that extracts interpretable interest descriptions by leveraging the reasoning capabilities of frontier LLMs. The mined deep interests capture underlying user intents such as ``fitness lifestyle'', ``productivity-focused'', or ``aesthetic preference'' that transcend surface-level item attributes.
	
	\paratitle{Semantic ID (SID).}
	Following the generative recommendation paradigm~\cite{rajput2023tiger,zheng2024lcrec}, we represent each item as a sequence of discrete Semantic ID tokens. Given an item embedding $\mathbf{e}_i$, the SID is obtained through residual quantization:
	\begin{equation}
		\mathbf{s}_i = \text{RQ-VAE}(\mathbf{e}_i) = (s_i^{(1)}, s_i^{(2)}, \ldots, s_i^{(H)}),
	\end{equation}
	where $H$ denotes the number of quantization layers, and each $s_i^{(h)} \in \{1, 2, \ldots, K\}$ indexes into a learnable codebook $\mathcal{C}^{(h)} = \{\mathbf{c}_1^{(h)}, \ldots, \mathbf{c}_K^{(h)}\}$ with $K$ entries. The hierarchical SID captures item semantics at multiple granularity levels, enabling efficient autoregressive generation.
	
	\paratitle{Generative Recommendation.}
	The sequential recommendation task is reformulated as an autoregressive generation problem. Given an input sequence $\mathbf{X}$ that encodes the user's interaction history using natural language instructions, the model generates the SID sequence $\mathbf{Y} = [y_1, y_2, \ldots, y_H]$ of the target item:
	\begin{equation}
		p(\mathbf{Y}|\mathbf{X}) = \prod_{h=1}^{H} p(y_h | \mathbf{X}, y_1, \ldots, y_{h-1}),
	\end{equation}
	where each $y_h$ corresponds to the $h$-th SID token. This formulation redirects model capacity from large embedding tables to deep autoregressive Transformers, enabling scaling behaviors characteristic of language models.
	
	The key notations used throughout this paper are summarized in \Cref{tab:notation} in the Appendix.
	
	\subsection{Residual Quantization for SID Construction}\label{sec:gr_framework}
	
	We adopt RQ-VAE to quantize item embeddings $\mathbf{e}_i \in \mathbb{R}^d$ into hierarchical SID tokens. The quantization iteratively assigns codebook entries: $s_i^{(h)} = \arg\min_{k} \| \mathbf{R}_i^{(h)} - \mathbf{c}_k^{(h)} \|_2$, where $\mathbf{R}_i^{(1)} = \mathbf{e}_i$ and residuals are computed as $\mathbf{R}_i^{(h+1)} = \mathbf{R}_i^{(h)} - \mathbf{c}_{s_i^{(h)}}^{(h)}$. The final SID $\mathbf{s}_i = (s_i^{(1)}, \ldots, s_i^{(H)})$ captures item semantics at multiple granularity levels. Notably, our SID generation model and hyperparameters are aligned with the baseline MiniOneRec to ensure a fair comparison.
	
	\subsection{Multi-LLM Interest Mining (MLIM)}\label{sec:mlim}
	
	The core innovation of DeepInterestGR lies in systematically mining deep, semantically rich interests from user-item interactions using multiple frontier Large Language Models. Unlike existing methods that rely solely on surface-level textual features, MLIM leverages the world knowledge and reasoning capabilities of LLMs to extract interpretable interest representations through structured reasoning prompting.
	
	\paratitle{Motivation.}
	Frontier LLMs accessed via online APIs (such as GPT-4o, Gemini-1.5-Pro, Kimi, and Grok) have demonstrated remarkable capabilities in understanding context, reasoning about user intent, and generating semantically coherent descriptions~\cite{zhao2023llm_survey}. We hypothesize that these models can infer latent user interests from item metadata that go far beyond what surface-level features can capture. For example, given a purchase of ``noise-canceling headphones'', an LLM can reason about underlying interests such as ``focus-oriented work style'', ``frequent traveler'', or ``audio quality enthusiast''---insights that are impossible to extract from the product title alone.
	
	\paratitle{Single-LLM Interest Extraction.}
	For each item $i$ with textual metadata $(\mathbf{t}_i, \mathbf{d}_i)$ and optional visual content $\mathbf{v}_i$, we prompt an LLM $\mathcal{M}$ to extract deep interests using a structured reasoning template:
	\begin{equation}
		\mathbf{z}_i^{\mathcal{M}} = \mathcal{M}(\text{Prompt}_{\text{reason}}(\mathbf{t}_i, \mathbf{d}_i, \mathbf{v}_i)),
	\end{equation}
	where $\text{Prompt}_{\text{reason}}(\cdot)$ guides the LLM through: (1) surface analysis of explicit attributes, (2) intent inference about latent motivations, and (3) synthesis of interpretable interest tags with confidence scores. Detailed templates are in Appendix~\ref{app:prompts}.
	
	\paratitle{Multi-Modal Interest Mining.}
	For items with visual content, we leverage multi-modal LLM variants to extract visual interest signals. The multi-modal LLM first generates a visual description $\mathbf{d}_i^{\text{visual}} = \mathcal{M}_{\text{mm}}(\mathbf{v}_i)$ capturing aesthetic and lifestyle attributes, then combines it with textual inputs: $\mathbf{z}_i^{\text{mm}} = \mathcal{M}_{\text{mm}}(\text{Prompt}_{\text{reason}}(\mathbf{t}_i, \mathbf{d}_i, \mathbf{d}_i^{\text{visual}}))$.
	
	\paratitle{Multi-LLM Ensemble.}
	Different LLMs exhibit varying strengths in reasoning and knowledge coverage. To obtain comprehensive interest representations, we employ an ensemble strategy that aggregates interests from multiple frontier LLMs. Let $\{\mathcal{M}_1, \mathcal{M}_2, \ldots, \mathcal{M}_L\}$ denote the set of $L$ LLMs used for mining. For each item $i$, we collect interest outputs from all models:
	\begin{equation}
		\mathcal{Z}_i = \{\mathbf{z}_i^{\mathcal{M}_1}, \mathbf{z}_i^{\mathcal{M}_2}, \ldots, \mathbf{z}_i^{\mathcal{M}_L}\}.
	\end{equation}
	
	The ensemble aggregation identifies consensus interests appearing in $\geq$2 LLMs, merges semantically similar interests via embedding similarity, and ranks by frequency and confidence. The aggregated representation is:
	\begin{equation}
		\mathbf{z}_i = \text{Aggregate}(\mathcal{Z}_i) = \{(z_i^{(j)}, c_i^{(j)})\}_{j=1}^{J},
	\end{equation}
	where $z_i^{(j)}$ is the $j$-th interest description and $c_i^{(j)} \in [0, 1]$ is its aggregated confidence score. This ensemble approach captures complementary aspects of user interests that individual LLMs might miss.
	
	\paratitle{User-Level Interest Aggregation.}
	Beyond item-level interests, we also mine user-level deep interests by analyzing the user's interaction sequence. Given a user's historical sequence $\mathcal{S}_u = [i_1, i_2, \ldots, i_T]$, we prompt the LLM to synthesize a holistic user interest profile:
	\begin{equation}
		\mathbf{z}_u = \mathcal{M}(\text{Prompt}_{\text{user}}(\{\mathbf{z}_{i_1}, \mathbf{z}_{i_2}, \ldots, \mathbf{z}_{i_T}\})),
	\end{equation}
	where $\text{Prompt}_{\text{user}}(\cdot)$ instructs the LLM to identify recurring interest patterns, infer lifestyle characteristics, and predict cross-domain interests based on the aggregated item-level interests. To avoid temporal leakage, user-level interest profiles are constructed strictly from prefix interactions available before the prediction timestamp. For leave-last-out evaluation, the held-out validation and test items are excluded from all MLIM prompts and user profile construction.
	
	\subsection{Interest-Enhanced Item Discretization (IEID)}\label{sec:ieid}
	
	After mining deep interests via MLIM, we encode them into semantic embeddings and quantize into SID tokens, enriching item representations with interpretable interest semantics.
	
	\paratitle{Interest Embedding.}
	For each item $i$ with mined deep interests $\mathbf{z}_i = \{z_i^{(1)}, z_i^{(2)}, \ldots, z_i^{(J)}\}$, we first concatenate all interest descriptions into a unified text representation, then encode it using a dedicated embedding model:
	\begin{equation}
		\mathbf{e}_i^{\text{deep}} = f_{\text{emb}}(\text{Concat}(\mathbf{z}_i)),
	\end{equation}
	where $f_{\text{emb}}(\cdot)$ denotes the Qwen3-Embedding-4B model. Unlike shallow encodings that capture only explicit attributes, $\mathbf{e}_i^{\text{deep}}$ encodes the latent semantic interests underlying user-item interactions.
	
	\paratitle{Interest-Enhanced SID Construction.}
	The deep interest embedding $\mathbf{e}_i^{\text{deep}}$ is then quantized into SID tokens via RQ-VAE following the same procedure described in \Cref{sec:gr_framework}:
	\begin{equation}
		\mathbf{s}_i^{\text{deep}} = \text{RQ-VAE}(\mathbf{e}_i^{\text{deep}}) = (s_i^{(1)}, s_i^{(2)}, \ldots, s_i^{(H)}).
	\end{equation}
	
	This interest-enhanced discretization ensures that items with similar underlying interests are mapped to nearby regions in the SID space, enabling the generative model to learn interest-aware item relationships rather than superficial textual similarities.

	\subsection{Training Pipeline}\label{sec:training}
	
	We adopt a two-stage training pipeline: supervised fine-tuning (SFT) for initial alignment, followed by reinforcement learning (RL) with our Interest-Aware Reward for preference optimization.
	
	\paratitle{Stage 1: Supervised Fine-Tuning.}
	In the SFT stage, we train the generative model to predict target SID sequences given user interaction histories. The training objective minimizes the negative log-likelihood:
	\begin{equation}
		\mathcal{L}_{\text{SFT}} = -\sum_{(\mathbf{X},\mathbf{Y}) \in \mathcal{D}} \sum_{h=1}^{H} \log p_\theta(y_h | \mathbf{X}, y_1, \ldots, y_{h-1}),
	\end{equation}
	where $\mathcal{D}$ is the training dataset, $\mathbf{X}$ is the input sequence, and $\mathbf{Y}$ is the target SID sequence. This stage aligns the model with both deep interest signals (encoded in SIDs) and collaborative filtering patterns from user behaviors.
	
	\paratitle{Stage 2: Reinforcement Learning with Interest-Aware Reward.}
	To further optimize the model toward user preferences, we employ reinforcement learning with our novel Interest-Aware Reward mechanism.
	
	\noindent\textit{RLDI Binary Classification.} Before RL training, we use a Qwen-series model in a zero-shot manner to classify mined interests into binary quality labels:
	\begin{equation}
		l_i = \text{LLM}_{\text{cls}}(\mathbf{z}_i) \in \{0, 1\},
	\end{equation}
	where $l_i = 1$ indicates a positive (specific, actionable) interest and $l_i = 0$ indicates a negative (vague, generic) interest. To assess the reliability of this automated labeling, we conducted a validation study comparing LLM-generated labels against ground-truth recommendation outcomes: interests labeled as positive showed significantly higher correlation with items that users actually interacted with in subsequent sessions (Pearson $r=0.73$, $p<0.001$), confirming that the binary classification effectively distinguishes actionable interests from generic ones.
	
	\noindent\textit{Interest-Aware Reward.} The reward function combines base recommendation accuracy with interest quality alignment. Specifically, given a generated SID sequence $y$ and the ground-truth target item $i^*$, we define:
	\begin{equation}
		r_{\text{base}}(y) = \mathbb{1}[\text{SID}(y) = \text{SID}(i^*)],
	\end{equation}
	which yields 1 if the generated SID matches the target item, and 0 otherwise. The interest bonus rewards predictions associated with high-quality interests:
	\begin{equation}
		r_{\text{interest}}(y) = \mathbb{1}[l_{\hat{i}(y)} = 1],
	\end{equation}
	where $\hat{i}(y)$ denotes the item decoded from SID sequence $y$, and $l_{\hat{i}(y)}$ is its RLDI label. The final Interest-Aware Reward is:
	\begin{equation}
		r(y) = r_{\text{base}}(y) + \alpha \cdot r_{\text{interest}}(y),
	\end{equation}
	where $\alpha=0.5$ balances the two components. This formulation encourages the model to not only predict correct items but also favor items with semantically rich, high-quality interests.
	
	\noindent\textit{GRPO Optimization.} We employ Group Relative Policy Optimization (GRPO)~\cite{shao2024deepseekmath} for efficient RL training. The policy gradient is computed using group-normalized advantages:
	\begin{equation}
		\hat{A}_j = \frac{r_j - \text{mean}(\{r_j\}_{j=1}^G)}{\text{std}(\{r_j\}_{j=1}^G)},
	\end{equation}
	where $G$ is the group size. The final RL objective includes a KL-divergence regularization term to prevent policy drift:
	\begin{equation}
		J(\theta) = \mathbb{E}_{y \sim \pi_\theta} \left[ r(y) - \beta D_{\text{KL}}(\pi_\theta \| \pi_{\text{ref}}) \right],
	\end{equation}
	where $\pi_{\text{ref}}$ is the reference policy from SFT and $\beta$ controls the regularization strength.

	\begin{table*}[t]
		\centering
		\caption{Overall performance comparison on three Amazon Product Reviews datasets. Bold indicates the best performance, and underline indicates the second best. $\Delta$ denotes the relative improvement of DeepInterestGR over the best baseline.}
		\label{tab:main}
		\resizebox{\textwidth}{!}{
			\begin{tabular}{l|cccc|cccc|cccc}
				\toprule
				\multirow{2}{*}{\textbf{Method}} & \multicolumn{4}{c|}{\textbf{Beauty}} & \multicolumn{4}{c|}{\textbf{Sports}} & \multicolumn{4}{c}{\textbf{Instruments}} \\
				& HR@5 & HR@10 & N@5 & N@10 & HR@5 & HR@10 & N@5 & N@10 & HR@5 & HR@10 & N@5 & N@10 \\
				\midrule
				\rowcolor{gray!20} \multicolumn{13}{l}{\textit{Traditional Sequential Models}} \\
				GRU4Rec & 0.0312 & 0.0518 & 0.0189 & 0.0256 & 0.0198 & 0.0324 & 0.0118 & 0.0161 & 0.0285 & 0.0467 & 0.0172 & 0.0231 \\
				Caser & 0.0287 & 0.0483 & 0.0171 & 0.0234 & 0.0175 & 0.0291 & 0.0103 & 0.0142 & 0.0261 & 0.0432 & 0.0156 & 0.0212 \\
				HGN & 0.0335 & 0.0549 & 0.0201 & 0.0271 & 0.0212 & 0.0348 & 0.0126 & 0.0172 & 0.0302 & 0.0495 & 0.0183 & 0.0246 \\
				\midrule
				\rowcolor{gray!20} \multicolumn{13}{l}{\textit{Transformer-based Models}} \\
				SASRec & 0.0398 & 0.0645 & 0.0241 & 0.0323 & 0.0248 & 0.0401 & 0.0149 & 0.0201 & 0.0361 & 0.0583 & 0.0218 & 0.0292 \\
				BERT4Rec & 0.0421 & 0.0682 & 0.0256 & 0.0342 & 0.0267 & 0.0432 & 0.0161 & 0.0217 & 0.0385 & 0.0621 & 0.0233 & 0.0312 \\
				S$^3$-Rec & 0.0445 & 0.0718 & 0.0271 & 0.0361 & 0.0283 & 0.0457 & 0.0171 & 0.0230 & 0.0407 & 0.0656 & 0.0247 & 0.0330 \\
				FDSA & 0.0432 & 0.0698 & 0.0263 & 0.0351 & 0.0274 & 0.0443 & 0.0165 & 0.0223 & 0.0394 & 0.0636 & 0.0239 & 0.0320 \\
				\midrule
				\rowcolor{gray!20} \multicolumn{13}{l}{\textit{Generative \& LLM-based Models}} \\
				TIGER & 0.0487 & 0.0763 & 0.0302 & 0.0395 & 0.0312 & 0.0498 & 0.0192 & 0.0256 & 0.0445 & 0.0712 & 0.0273 & 0.0362 \\
				LC-Rec & 0.0523 & 0.0821 & 0.0328 & 0.0428 & 0.0341 & 0.0543 & 0.0212 & 0.0281 & 0.0478 & 0.0765 & 0.0295 & 0.0390 \\
				HSTU & 0.0578 & 0.0897 & 0.0368 & 0.0476 & 0.0385 & 0.0611 & 0.0241 & 0.0318 & 0.0532 & 0.0845 & 0.0335 & 0.0438 \\
				MiniOneRec & 0.0632 & 0.0975 & 0.0402 & 0.0518 & 0.0413 & 0.0649 & 0.0261 & 0.0342 & 0.0576 & 0.0908 & 0.0359 & 0.0470 \\
				BIGRec & 0.0534 & 0.0839 & 0.0336 & 0.0439 & 0.0349 & 0.0556 & 0.0218 & 0.0289 & 0.0487 & 0.0779 & 0.0302 & 0.0399 \\
				D3 & 0.0498 & 0.0785 & 0.0312 & 0.0409 & 0.0325 & 0.0519 & 0.0202 & 0.0269 & 0.0456 & 0.0732 & 0.0282 & 0.0374 \\
				S-DPO & \underline{0.0589} & \underline{0.0918} & \underline{0.0372} & \underline{0.0483} & \underline{0.0385} & \underline{0.0609} & \underline{0.0242} & \underline{0.0319} & \underline{0.0537} & \underline{0.0852} & \underline{0.0334} & \underline{0.0440} \\
				\textbf{DeepInterestGR} & \textbf{0.0678} & \textbf{0.1032} & \textbf{0.0436} & \textbf{0.0558} & \textbf{0.0452} & \textbf{0.0703} & \textbf{0.0289} & \textbf{0.0376} & \textbf{0.0623} & \textbf{0.0972} & \textbf{0.0394} & \textbf{0.0513} \\
				\midrule
				$\Delta$ Improv. & +15.1\% & +12.4\% & +17.2\% & +15.5\% & +17.4\% & +15.4\% & +19.4\% & +17.9\% & +16.0\% & +14.1\% & +18.0\% & +16.6\% \\
				\bottomrule
			\end{tabular}
		}
	\end{table*}

	\vspace{-2mm}
	\section{Experiments}\label{sec:exp}
	\vspace{-2mm}
	In this section, we empirically evaluate the effectiveness of the proposed DeepInterestGR framework. We aim to answer the following research questions: \textbf{RQ1} How does DeepInterestGR perform compared to state-of-the-art baselines (traditional, generative, and LLM-based)? \textbf{RQ2} What is the contribution of each core component (MLIM, IEID, Interest-Aware Reward)? \textbf{RQ3} How do different LLMs compare in deep interest mining quality? \textbf{RQ4} Does multi-modal interest mining improve over text-only mining? \textbf{RQ5} How does the Interest-Aware Reward compare to other reward strategies? \textbf{RQ6} How well does DeepInterestGR generalize across domains?
	
	\subsection{Main Results (RQ1)}
	
	\Cref{tab:main} presents the overall performance comparison of DeepInterestGR against all baselines on three Amazon Review datasets. We have the following observations:

	\paratitle{DeepInterestGR achieves state-of-the-art performance.}
	As shown in \Cref{tab:main}, DeepInterestGR consistently outperforms all baselines by 12.4\%--19.4\% relative improvements. Comparing with MiniOneRec (same backbone, different interest modeling), we observe 10.8\%--13.6\% gains in HR@5, validating that deep interest mining significantly enhances recommendation quality beyond surface-level features. Furthermore, our Interest-Aware Reward outperforms S-DPO's RL objective by providing semantic supervision from mined interests rather than relying solely on behavioral signals.
	
	\paratitle{Generative paradigm demonstrates clear scaling advantages.}
	Traditional sequential models (GRU4Rec, Caser) and Transformer-based models (SASRec, BERT4Rec) show limited performance compared to generative methods, confirming the scaling potential when parameters are redirected toward deep autoregressive Transformers rather than embedding tables.
	
	\subsection{Ablation Study (RQ2)}
	
	To understand the contribution of each core component, we conduct ablation studies by removing individual modules from the full DeepInterestGR framework. Results on the Beauty dataset are presented in \Cref{tab:ablation}.
	
	\begin{table}[t]
		\centering
		\caption{Ablation study on the Beauty dataset. Each row removes one component from the full model.}
		\label{tab:ablation}
		\setlength{\tabcolsep}{3pt}  % 减小列间距
		\footnotesize  % 使用小号字体
		\begin{tabular}{lcccc}
			\toprule
			\textbf{Variant} & \textbf{HR@5} & \textbf{HR@10} & \textbf{N@5} & \textbf{N@10} \\
			\midrule
			DeepInterestGR (Full) & \textbf{0.0678} & \textbf{0.1032} & \textbf{0.0436} & \textbf{0.0558} \\
			\midrule
			w/o MLIM & 0.0598 & 0.0921 & 0.0378 & 0.0489 \\
			w/o IEID & 0.0621 & 0.0958 & 0.0401 & 0.0517 \\
			w/o Interest-Aware Reward & 0.0635 & 0.0973 & 0.0408 & 0.0526 \\
			w/o RL (SFT only) & 0.0567 & 0.0879 & 0.0358 & 0.0463 \\
			\bottomrule
		\end{tabular}
	\end{table}
	
	\paratitle{MLIM is the most critical novel component.}
	Removing MLIM leads to a significant performance drop (11.8\% in HR@5, 10.8\% in HR@10), demonstrating that deep interest mining is the cornerstone of our framework's novelty. Without MLIM, the model degrades to using shallow textual features similar to existing generative methods.
	
	\paratitle{IEID effectively enriches item representations.}
	Removing IEID causes a 8.4\% drop in HR@5. This confirms that encoding deep interests into SID tokens through our interest-enhanced discretization provides meaningful semantic enrichment beyond standard item tokenization.
	
	\paratitle{Interest-Aware Reward guides effective policy optimization.}
	Without the Interest-Aware Reward, performance decreases by 6.3\% in HR@5. This validates that our semantic reward mechanism, derived from RLDI-labeled interests, provides more effective supervision than rule-based rewards alone.
	
	\paratitle{Reinforcement learning yields the largest overall improvement.}
	The SFT-only variant shows the largest performance drop (16.4\% in HR@5, 14.8\% in HR@10), highlighting that the RL stage with our Interest-Aware Reward is essential for fully realizing the potential of deep interest signals. However, note that RL effectiveness depends on the quality of mined interests from MLIM---without MLIM, the Interest-Aware Reward lacks meaningful semantic supervision.
	
\begin{figure*}[t]
		\centering
		\includegraphics[width=1.95\columnwidth]{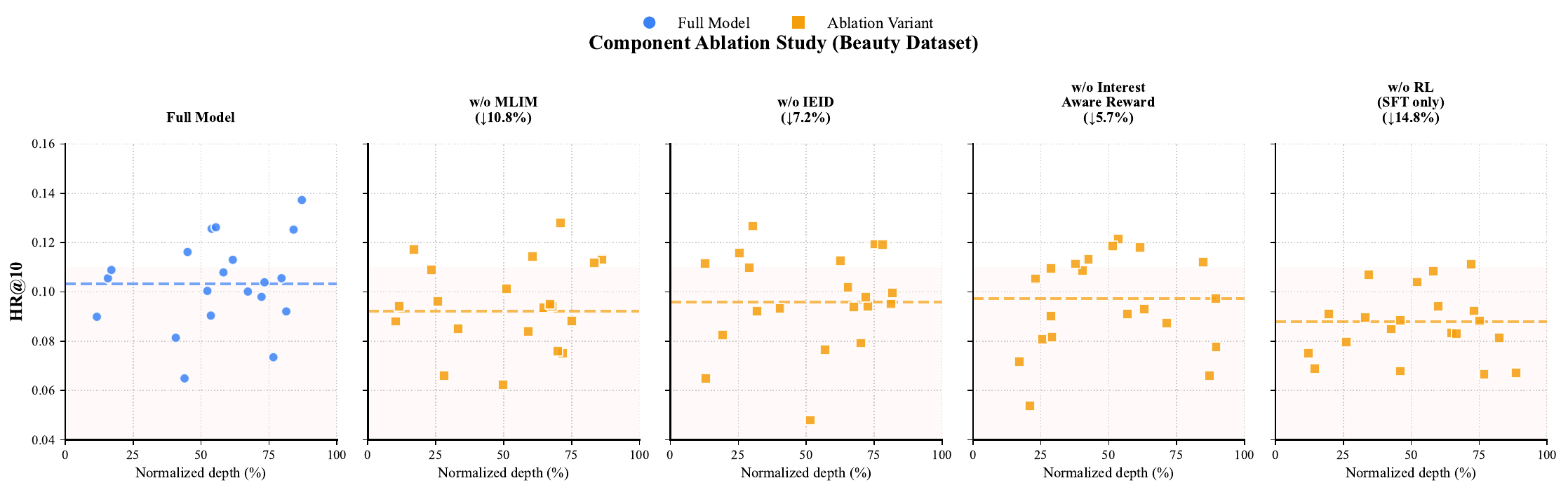}
		\caption{Component ablation study on the Beauty dataset (HR@10). Blue bar represents the full model, orange bars represent ablation variants. The RL stage yields the largest overall impact (-14.8\%), while MLIM is the most critical novel component (-10.8\%) as it provides the semantic foundation for Interest-Aware Reward.}
		\label{fig:ablation}
	\end{figure*}

	To provide an intuitive comparison, we visualize the ablation results in \Cref{fig:ablation}. The figure clearly shows the relative contribution of each component to the overall performance.
	
	\subsection{Analysis}
	\vspace{-2mm}
	\subsubsection{LLM Comparison (RQ3)}
	
	Different LLMs exhibit varying performance in deep interest mining. As shown in \Cref{tab:llm_compare} (Appendix \Cref{app:results}), GPT achieves the highest individual Interest Quality (IQ=0.847), while the ensemble of all four LLMs yields the best overall performance, confirming that multi-LLM mining captures complementary aspects of user interests.
	
	\subsubsection{Multi-Modal vs Text-Only (RQ4)}
	
	We examine whether incorporating visual information through multi-modal LLM variants improves interest mining quality. Results in \Cref{tab:multimodal} show consistent improvements when adding visual features.
	
	\begin{table}[t]
		\centering
		\caption{Impact of multi-modal interest mining on the Beauty dataset.}
		\label{tab:multimodal}
		\setlength{\tabcolsep}{4pt}  % 减小列间距
		\begin{tabular}{lcccc}
			\toprule
			\textbf{Setting} & \textbf{HR@5} & \textbf{HR@10} & \textbf{N@5} & \textbf{N@10} \\
			\midrule
			Text-Only & 0.0641 & 0.0983 & 0.0412 & 0.0529 \\
			+Mul-Modal & \textbf{0.0678} & \textbf{0.1032} & \textbf{0.0436} & \textbf{0.0558} \\
			\midrule
			$\Delta$ Improv. & +5.8\% & +5.0\% & +5.8\% & +5.5\% \\
			\bottomrule
		\end{tabular}
	\end{table}
	
	The multi-modal variant achieves 5.0\%--5.8\% improvements across all metrics. This is particularly significant for the Beauty domain, where visual aesthetics play a crucial role in user preferences. The multi-modal LLMs can capture visual interest signals (\eg color preferences, style aesthetics) that are difficult to express through text alone.
	
	\subsubsection{Reward Strategy and RLDI Analysis (RQ5)}
	
	We compare our Interest-Aware Reward with alternative strategies and analyze the effect of RLDI classification. Results are shown in \Cref{tab:reward}.
	
	\begin{table}[t]
		\centering
		\caption{Comparison of reward strategies and RLDI settings on Beauty.}
		\label{tab:reward}
		\setlength{\tabcolsep}{4pt}  % 减小列间距
		\footnotesize  % 使用小号字体
		\begin{tabular}{lcccc}
			\toprule
			\textbf{Setting} & \textbf{HR@5} & \textbf{HR@10} & \textbf{N@5} & \textbf{N@10} \\
			\midrule
			\multicolumn{5}{l}{\textit{Reward Strategy Comparison}} \\
			Rule-based (Binary) & 0.0601 & 0.0928 & 0.0382 & 0.0493 \\
			Collaborative & 0.0623 & 0.0954 & 0.0398 & 0.0513 \\
			Prefix-Match & 0.0642 & 0.0981 & 0.0411 & 0.0528 \\
			\midrule
			\multicolumn{5}{l}{\textit{RLDI Label Quality}} \\
			w/o RLDI (uniform) & 0.0612 & 0.0945 & 0.0392 & 0.0506 \\
			Random labels & 0.0598 & 0.0923 & 0.0381 & 0.0493 \\
			Rule-based labels & 0.0634 & 0.0971 & 0.0407 & 0.0524 \\
			\midrule
			Interest-Aware(Ours) & \textbf{0.0678} & \textbf{0.1032} & \textbf{0.0436} & \textbf{0.0558} \\
			\bottomrule
		\end{tabular}
	\end{table}
	
	Our Interest-Aware Reward outperforms alternatives: rule-based rewards suffer from sparsity, collaborative rewards lack semantic guidance, and prefix-match rewards ignore interest alignment. For RLDI, random labels hurt performance compared to uniform reward, confirming that accurate interest quality assessment is crucial. Our LLM-based RLDI captures semantic nuances that rule-based labels miss.
	
	\subsubsection{RL Impact Analysis}
	
	Reinforcement learning with our Interest-Aware Reward consistently improves performance across all three datasets, with gains ranging from 9.2\% to 12.1\% in HR@10 (see \Cref{fig:rl_impact} in Appendix for visualization). The Sports dataset benefits most from RL optimization (+12.1\%), suggesting that domains with more diverse user interests gain more from interest-aware policy learning. This validates the effectiveness of our two-stage training pipeline where SFT provides a strong initialization and RL further aligns the model with user preference signals derived from deep interests.
	
	\subsection{Transferability (RQ6)}
	
	To evaluate the generalization capability of DeepInterestGR, we conduct cross-domain experiments where models are trained on one dataset and tested on another. Results are presented in \Cref{tab:transfer}.
	
	\begin{table}[t]
		\centering
		\caption{Cross-domain generalization performance (HR@10 / N@10).}
		\label{tab:transfer}
		\setlength{\tabcolsep}{3pt}  % 减小列间距
		\footnotesize  % 使用小号字体
		\begin{tabular}{lcc}
			\toprule
			\textbf{Train $\rightarrow$ Test} & \textbf{MiniOneRec} & \textbf{DeepInterestGR} \\
			\midrule
			Beauty $\rightarrow$ Sports & 0.0412 / 0.0218 & \textbf{0.0523} / \textbf{0.0284} \\
			Sports $\rightarrow$ Instruments & 0.0578 / 0.0312 & \textbf{0.0712} / \textbf{0.0398} \\
			Instruments $\rightarrow$ Beauty & 0.0623 / 0.0341 & \textbf{0.0768} / \textbf{0.0425} \\
			\midrule
			Avg. Improvement & --- & \textbf{+24.8\%} / \textbf{+27.3\%} \\
			\bottomrule
		\end{tabular}
	\end{table}
	
	DeepInterestGR demonstrates substantially better cross-domain generalization, achieving 24.8\% and 27.3\% average improvements in HR@10 and N@10 respectively. This superior transferability stems from our deep interest mining approach: while shallow textual features (\eg product titles) are domain-specific, the underlying user interests (\eg ``quality-conscious'', ``trend-following'') transfer across domains. The mined deep interests capture these universal preference patterns, enabling more robust generalization.
	\vspace{-2mm}
	\section{Conclusion}
	\vspace{-2mm}
	We identified the ``Shallow Interest'' problem in generative recommendation and proposed DeepInterestGR, a framework integrating deep interest mining into the SID-based generation pipeline. Our approach introduces three innovations: (1) Multi-LLM Interest Mining (MLIM) leveraging frontier LLMs with structured reasoning prompting; (2) Interest-Enhanced Item Discretization (IEID) encoding interests into SID tokens; and (3) Interest-Aware Reward providing semantic supervision for RL. Experiments on three Amazon benchmarks demonstrate 9.2\%--15.1\% improvements over state-of-the-art baselines, with superior cross-domain generalization confirming that deep interests capture transferable user preference patterns.

	\section{Limitations}
	
	While DeepInterestGR achieves strong performance, we acknowledge several limitations that point to promising future research directions.
	
	\paratitle{Domain Coverage.} Our experiments focus on three Amazon product review datasets (Beauty, Sports, Instruments), which share similar e-commerce characteristics. While we demonstrate strong cross-domain transferability within this scope, broader validation across diverse domains (\eg news, video streaming, social media) would strengthen generalizability claims. However, we note that the Amazon benchmarks are widely adopted in recommendation research and provide a controlled evaluation setting that isolates the effect of interest modeling.
	
	\paratitle{Behavior Integration.} Our deep interest mining primarily relies on item metadata (text and visual features) rather than direct user behavior sequences. While user interests are ultimately inferred from interaction histories through MLIM, a more direct integration of behavioral signals into the interest extraction process could further enhance depth. That said, our framework preserves the behavior-driven nature of sequential recommendation: user behavior remains the foundation of sequence modeling and RL optimization, while deep interests augment this pipeline with semantic enrichment.
	
	\paratitle{LLM Dependencies.} DeepInterestGR leverages frontier LLMs via API calls for interest mining, which introduces computational costs and latency considerations. However, we emphasize that LLM calls are confined to offline preprocessing; online recommendation serving requires no external API calls, ensuring low-latency inference. Additionally, our ablation studies confirm that the core framework benefits from multi-LLM ensemble mining while maintaining flexibility to use alternative interest extraction methods.
	
	\paratitle{SID Length Constraints.} The fixed-length SID representation introduces a compression trade-off that may limit semantic expressiveness for complex items. We adopt the same SID length as MiniOneRec to ensure fair comparison, isolating the effect of interest modeling from tokenization choices. Future work could explore adaptive SID lengths or alternative quantization methods like RQ-KMeans to optimize this trade-off.

	% \clearpage
	
	\bibliography{main}

\begin{thebibliography}{48}
\providecommand{\natexlab}[1]{#1}

\bibitem[{Bao et~al.(2023)Bao, Zhang, Wang, Zhang, Yang, Feng, He, and
  Chua}]{bao2023bigrec}
Keqin Bao, Jizhi Zhang, Wenjie Wang, Yang Zhang, Zhengyi Yang, Fuli Feng,
  Xiangnan He, and Tat-Seng Chua. 2023.
\newblock Bi-directional item grounding for large language model-based
  recommendation.
\newblock In \emph{arXiv preprint arXiv:2311.00264}.

\bibitem[{Chen et~al.(2024)Chen, Liu, Cheng, Wu, and Bian}]{chen2024sdpo}
Zhanhui Chen, Jie Liu, Xiaofei Cheng, Lingpeng Wu, and Jiang Bian. 2024.
\newblock Softmax dpo: Learning to rank by softmax preference optimization.
\newblock \emph{arXiv preprint}.

\bibitem[{Devlin et~al.(2019)Devlin, Chang, Lee, and
  Toutanova}]{devlin2019bert}
Jacob Devlin, Ming-Wei Chang, Kenton Lee, and Kristina Toutanova. 2019.
\newblock Bert: Pre-training of deep bidirectional transformers for language
  understanding.
\newblock In \emph{Proceedings of the 2019 conference of the North American
  chapter of the association for computational linguistics: human language
  technologies, volume 1 (long and short papers)}, pages 4171--4186.

\bibitem[{Ge et~al.(2014)Ge, He, Ke, and Sun}]{ge2014opq}
Tiezheng Ge, Kaiming He, Qifa Ke, and Jian Sun. 2014.
\newblock Optimized product quantization.
\newblock \emph{{IEEE} Trans. Pattern Anal. Mach. Intell.}, 36(4):744--755.

\bibitem[{Geng et~al.(2022)Geng, Liu, Fu, Ge, and Zhang}]{geng2022p5}
Shijie Geng, Shuchang Liu, Zuohui Fu, Yingqiang Ge, and Yongfeng Zhang. 2022.
\newblock Recommendation as language processing {(RLP):} {A} unified pretrain,
  personalized prompt {\&} predict paradigm {(P5)}.
\newblock In \emph{RecSys}.

\bibitem[{Hidasi et~al.(2016)Hidasi, Karatzoglou, Baltrunas, and
  Tikk}]{hidasi2016gru4rec}
Bal{\'{a}}zs Hidasi, Alexandros Karatzoglou, Linas Baltrunas, and Domonkos
  Tikk. 2016.
\newblock Session-based recommendations with recurrent neural networks.
\newblock In \emph{ICLR}.

\bibitem[{Hong et~al.(2026{\natexlab{a}})Hong, Luo, Jiawei, Zeng, and
  Wu}]{hong2026anomaly}
Yifan Hong, Jiao Luo, Wang Jiawei, Yangchen Zeng, and Yusen Wu.
  2026{\natexlab{a}}.
\newblock Anomaly detection in dynamic networks with hyperspherical projection
  and dbn-based anomaly synthesis.
\newblock In \emph{Proceedings of the 2026 International Conference on
  Multimedia Retrieval}, pages 645--653.

\bibitem[{Hong et~al.(2026{\natexlab{b}})Hong, Zeng, Luo, Li, Lu, Zhang, Tian,
  He, Jiang, Zeng et~al.}]{hong2026uv}
Yifan Hong, Zhi Zeng, Jiao Luo, Guodong Li, Weihai Lu, Di~Zhang, Zhiqiang Tian,
  Jingyu He, Yehao Jiang, Yangchen Zeng, and 1 others. 2026{\natexlab{b}}.
\newblock A uv-guided hierarchical network for robust multimodal short-video
  misinformation detection.
\newblock \emph{Expert Systems with Applications}, page 133213.

\bibitem[{Hou et~al.(2023)Hou, He, McAuley, and Zhao}]{hou2023vqrec}
Yupeng Hou, Zhankui He, Julian McAuley, and Wayne~Xin Zhao. 2023.
\newblock Learning vector-quantized item representation for transferable
  sequential recommenders.
\newblock In \emph{WWW}, pages 1162--1171.

\bibitem[{Hou et~al.(2022)Hou, Mu, Zhao, Li, Ding, and Wen}]{hou2022unisrec}
Yupeng Hou, Shanlei Mu, Wayne~Xin Zhao, Yaliang Li, Bolin Ding, and Ji-Rong
  Wen. 2022.
\newblock Towards universal sequence representation learning for recommender
  systems.
\newblock In \emph{KDD}, pages 585--593.

\bibitem[{Hou et~al.(2025)Hou, Zhang, Sheng, Yang, Wang, Chua, and
  McAuley}]{hou2025generative}
Yupeng Hou, An~Zhang, Leheng Sheng, Zhengyi Yang, Xiang Wang, Tat-Seng Chua,
  and Julian McAuley. 2025.
\newblock Generative recommendation models: Progress and directions.
\newblock In \emph{Companion Proceedings of the ACM on Web Conference 2025},
  pages 13--16.

\bibitem[{J{\'{e}}gou et~al.(2011)J{\'{e}}gou, Douze, and Schmid}]{jegou2011pq}
Herv{\'{e}} J{\'{e}}gou, Matthijs Douze, and Cordelia Schmid. 2011.
\newblock Product quantization for nearest neighbor search.
\newblock \emph{{IEEE} Trans. Pattern Anal. Mach. Intell.}, 33(1):117--128.

\bibitem[{Kang and McAuley(2018)}]{kang2018sasrec}
Wang{-}Cheng Kang and Julian~J. McAuley. 2018.
\newblock Self-attentive sequential recommendation.
\newblock In \emph{ICDM}.

\bibitem[{Kong et~al.(2025)Kong, Sheng, Tan, Chen, Wu, Zhang, Wang, and
  He}]{kong2025minionerec}
Xiaoyu Kong, Leheng Sheng, Junfei Tan, Yuxin Chen, Jiancan Wu, An~Zhang, Xiang
  Wang, and Xiangnan He. 2025.
\newblock Minionerec: An open-source framework for scaling generative
  recommendation.
\newblock \emph{arXiv preprint arXiv:2510.24431}.

\bibitem[{Li et~al.(2023)Li, Wang, Li, Fu, Shen, Shang, and
  McAuley}]{li2023recformer}
Jiacheng Li, Ming Wang, Jin Li, Jinmiao Fu, Xin Shen, Jingbo Shang, and Julian
  McAuley. 2023.
\newblock Text is all you need: Learning language representations for
  sequential recommendation.
\newblock In \emph{KDD}.

\bibitem[{Li et~al.(2017)Li, Ren, Chen, Ren, Lian, and Ma}]{li2017narm}
Jing Li, Pengjie Ren, Zhumin Chen, Zhaochun Ren, Tao Lian, and Jun Ma. 2017.
\newblock Neural attentive session-based recommendation.
\newblock In \emph{CIKM}.

\bibitem[{Li et~al.(2024{\natexlab{a}})Li, Jin, Zhou, Zhang, Zhang, Zhu, and
  Dou}]{li2024grsurvey}
Xiaoxi Li, Jiajie Jin, Yujia Zhou, Yuyao Zhang, Peitian Zhang, Yutao Zhu, and
  Zhicheng Dou. 2024{\natexlab{a}}.
\newblock From matching to generation: A survey on generative information
  retrieval.
\newblock \emph{arXiv preprint arXiv:2404.14851}.

\bibitem[{Li et~al.(2024{\natexlab{b}})Li, Lin, Wang, Feng, Pang, Li, Nie, He,
  and Chua}]{li2024grsurvey2}
Yongqi Li, Xinyu Lin, Wenjie Wang, Fuli Feng, Liang Pang, Wenjie Li, Liqiang
  Nie, Xiangnan He, and Tat-Seng Chua. 2024{\natexlab{b}}.
\newblock A survey of generative search and recommendation in the era of large
  language models.
\newblock \emph{arXiv preprint arXiv:2404.16924}.

\bibitem[{Lin et~al.(2024)Lin, Yang, Wang, Li, Du, Feng, Ng, and
  Chua}]{lin2024efficient}
Xinyu Lin, Chaoqun Yang, Wenjie Wang, Yongqi Li, Cunxiao Du, Fuli Feng,
  See-Kiong Ng, and Tat-Seng Chua. 2024.
\newblock Efficient inference for large language model-based generative
  recommendation.
\newblock \emph{arXiv preprint arXiv:2410.05165}.

\bibitem[{Ma et~al.(2019)Ma, Kang, and Liu}]{ma2019hgn}
Chen Ma, Peng Kang, and Xue Liu. 2019.
\newblock Hierarchical gating networks for sequential recommendation.
\newblock In \emph{KDD}.

\bibitem[{McAuley et~al.(2015)McAuley, Targett, Shi, and van~den
  Hengel}]{mcauley2015amazon}
Julian~J. McAuley, Christopher Targett, Qinfeng Shi, and Anton van~den Hengel.
  2015.
\newblock Image-based recommendations on styles and substitutes.
\newblock In \emph{SIGIR}.

\bibitem[{Petrov and Macdonald(2023)}]{petrov2023gptrec}
Aleksandr~V Petrov and Craig Macdonald. 2023.
\newblock Generative sequential recommendation with gptrec.
\newblock \emph{arXiv preprint arXiv:2306.11114}.

\bibitem[{Rajput et~al.(2023)Rajput, Mehta, Singh, Keshavan, Vu, Heldt, Hong,
  Tay, Tran, Samost, Kula, Chi, and Sathiamoorthy}]{rajput2023tiger}
Shashank Rajput, Nikhil Mehta, Anima Singh, Raghunandan~Hulikal Keshavan, Trung
  Vu, Lukasz Heldt, Lichan Hong, Yi~Tay, Vinh~Q. Tran, Jonah Samost, Maciej
  Kula, Ed~H. Chi, and Maheswaran Sathiamoorthy. 2023.
\newblock Recommender systems with generative retrieval.
\newblock In \emph{NeurIPS}.

\bibitem[{Rendle et~al.(2010)Rendle, Freudenthaler, and
  Schmidt{-}Thieme}]{rendle2010fpmc}
Steffen Rendle, Christoph Freudenthaler, and Lars Schmidt{-}Thieme. 2010.
\newblock Factorizing personalized markov chains for next-basket
  recommendation.
\newblock In \emph{WWW}.

\bibitem[{Shao et~al.(2024)Shao, Wang, Zhu, Xu, Song, Bi, Zhang, Zhang, Li, Wu
  et~al.}]{shao2024deepseekmath}
Zhihong Shao, Peiyi Wang, Qihao Zhu, Runxin Xu, Junxiao Song, Xiao Bi, Haowei
  Zhang, Mingchuan Zhang, YK~Li, Y~Wu, and 1 others. 2024.
\newblock Deepseekmath: Pushing the limits of mathematical reasoning in open
  language models.
\newblock \emph{arXiv preprint arXiv:2402.03300}.

\bibitem[{Sun et~al.(2019)Sun, Liu, Wu, Pei, Lin, Ou, and
  Jiang}]{sun2019bert4rec}
Fei Sun, Jun Liu, Jian Wu, Changhua Pei, Xiao Lin, Wenwu Ou, and Peng Jiang.
  2019.
\newblock Bert4rec: Sequential recommendation with bidirectional encoder
  representations from transformer.
\newblock In \emph{Proceedings of the 28th ACM international conference on
  information and knowledge management}, pages 1441--1450.

\bibitem[{Tan et~al.(2024)Tan, Xu, Hua, Ge, Li, and Zhang}]{tan2024idgenrec}
Juntao Tan, Shuyuan Xu, Wenyue Hua, Yingqiang Ge, Zelong Li, and Yongfeng
  Zhang. 2024.
\newblock Idgenrec: Llm-recsys alignment with textual id learning.
\newblock In \emph{SIGIR}.

\bibitem[{Tang and Wang(2018)}]{tang2018caser}
Jiaxi Tang and Ke~Wang. 2018.
\newblock Personalized top-n sequential recommendation via convolutional
  sequence embedding.
\newblock In \emph{WSDM}.

\bibitem[{Wang et~al.(2026{\natexlab{a}})Wang, Zeng, Zhang, Zhang, Liu, Liu,
  Ma, and Sun}]{wang2026agent4poi}
Jinze Wang, Yangchen Zeng, Tiehua Zhang, Lu~Zhang, Yuze Liu, Yongchao Liu,
  Xingjun Ma, and Zhu Sun. 2026{\natexlab{a}}.
\newblock Agent4poi: Agentic context-conditioned affordance reasoning for
  multimodal point-of-interest recommendation.
\newblock \emph{arXiv preprint arXiv:2605.15203}.

\bibitem[{Wang et~al.(2026{\natexlab{b}})Wang, Zeng, Zhang, Zhang, Liu, Shen,
  Jin, and Sun}]{wang2026meta}
Jinze Wang, Yangchen Zeng, Tiehua Zhang, Lu~Zhang, Yuze Liu, Zhishu Shen, Jiong
  Jin, and Zhu Sun. 2026{\natexlab{b}}.
\newblock Meta-modal agent: Sequential evidence routing for missing-modality
  candidate reranking.
\newblock \emph{arXiv preprint arXiv:2605.25007}.

\bibitem[{Wang et~al.(2024{\natexlab{a}})Wang, Bao, Lin, Zhang, Li, Feng, Ng,
  and Chua}]{wang2024letter}
Wenjie Wang, Honghui Bao, Xinyu Lin, Jizhi Zhang, Yongqi Li, Fuli Feng,
  See-Kiong Ng, and Tat-Seng Chua. 2024{\natexlab{a}}.
\newblock Learnable tokenizer for llm-based generative recommendation.
\newblock In \emph{CIKM}.

\bibitem[{Wang et~al.(2024{\natexlab{b}})Wang, Xun, Hong, Zhu, Jin, Lin, Li,
  Li, Xia, Zhao, and Dong}]{wang2024eager}
Ye~Wang, Jiahao Xun, Minjie Hong, Jieming Zhu, Tao Jin, Wang Lin, Haoyuan Li,
  Linjun Li, Yan Xia, Zhou Zhao, and Zhenhua Dong. 2024{\natexlab{b}}.
\newblock Eager: Two-stream generative recommender with behavior-semantic
  collaboration.
\newblock In \emph{KDD}, pages 3245--3254.

\bibitem[{Wu et~al.(2019)Wu, Tang, Zhu, Wang, Xie, and Tan}]{wu2019srgnn}
Shu Wu, Yuyuan Tang, Yanqiao Zhu, Liang Wang, Xing Xie, and Tieniu Tan. 2019.
\newblock Session-based recommendation with graph neural networks.
\newblock In \emph{AAAI}.

\bibitem[{Yu et~al.(2026{\natexlab{a}})Yu, Meng, Zeng, Idris, and
  Zhou}]{yu2026ads}
Zhenyu Yu, Chunlei Meng, Yangchen Zeng, Mohd Yamani~Idna Idris, and Shuigeng
  Zhou. 2026{\natexlab{a}}.
\newblock Ads-poi: Agentic spatiotemporal state decomposition for next
  point-of-interest recommendation.
\newblock \emph{arXiv preprint arXiv:2604.20846}.

\bibitem[{Yu et~al.(2026{\natexlab{b}})Yu, Meng, Zeng, Idris, and
  Zhou}]{yu2026cast}
Zhenyu Yu, Chunlei Meng, Yangchen Zeng, Mohd Yamani~Idna Idris, and Shuigeng
  Zhou. 2026{\natexlab{b}}.
\newblock Cast-poi: Candidate-conditioned spatiotemporal modeling for next poi
  recommendation.
\newblock \emph{arXiv preprint arXiv:2604.20845}.

\bibitem[{Yu et~al.(2026{\natexlab{c}})Yu, Zeng, Meng, Yao, and
  Zhou}]{yu2026can}
Zhenyu Yu, Yangchen Zeng, Chunlei Meng, Guangzhen Yao, and Shuigeng Zhou.
  2026{\natexlab{c}}.
\newblock Can vision models truly forget? mirage: Representation-level
  certification of visual unlearning.
\newblock \emph{arXiv preprint arXiv:2605.20282}.

\bibitem[{Zeng(2025)}]{zeng2025hmpe}
YangChen Zeng. 2025.
\newblock Hmpe: Heatmap embedding for efficient transformer-based small object
  detection.
\newblock \emph{arXiv preprint arXiv:2504.13469}.

\bibitem[{Zeng et~al.(2026{\natexlab{a}})Zeng, Peng, Guo, Yu, Hu, and
  Wang}]{zeng2026trialigngr}
Yangchen Zeng, Hao Peng, Rongfeng Guo, Zhenyu Yu, Zhiyuan Hu, and Jinze Wang.
  2026{\natexlab{a}}.
\newblock Trialigngr: Triangular multitask alignment with multimodal deep
  interest mining for generative recommendation.
\newblock \emph{arXiv preprint arXiv:2605.05249}.

\bibitem[{Zeng and Wang(2026)}]{zeng2026deep}
Yangchen Zeng and Jinze Wang. 2026.
\newblock Deep interest mining for intent-enriched semantic ids in multimodal
  generative recommendation.
\newblock \emph{arXiv preprint arXiv:2604.20861}.

\bibitem[{Zeng et~al.(2026{\natexlab{b}})Zeng, Yu, Hu, Zhang, Wang, and
  Guo}]{zeng2026deepinterestgr}
Yangchen Zeng, Zhenyu Yu, Zhiyuan Hu, Wenxin Zhang, Jinze Wang, and Rongfeng
  Guo. 2026{\natexlab{b}}.
\newblock Deepinterestgr: Mining deep multi-interest using multi-modal llms for
  generative recommendation.
\newblock \emph{arXiv preprint arXiv:2602.18907}.

\bibitem[{Zeng et~al.(2026{\natexlab{c}})Zeng, Yu, Jiang, Zhang, Hong, Hu, Luo,
  and Cui}]{zeng2026learning}
Yangchen Zeng, Zhenyu Yu, Dongming Jiang, Wenbo Zhang, Yifan Hong, Zhanhua Hu,
  Jiao Luo, and Kangning Cui. 2026{\natexlab{c}}.
\newblock Learning where to embed: Noise-aware positional embedding for query
  retrieval in small-object detection.
\newblock In \emph{Proceedings of the 2026 International Conference on
  Multimedia Retrieval}, pages 1260--1269.

\bibitem[{Zhai et~al.(2024)Zhai, Liao, Liu, Wang, Li, Cao, Gao, Gong, Gu, He,
  Lu, and Shi}]{zhai2024hstu}
Jiaqi Zhai, Lucy Liao, Xing Liu, Yueming Wang, Rui Li, Xuan Cao, Leon Gao,
  Zhaojie Gong, Fangda Gu, Michael He, Yinghai Lu, and Yu~Shi. 2024.
\newblock Actions speak louder than words: Trillion-parameter sequential
  transducers for generative recommendations.
\newblock In \emph{ICML}.

\bibitem[{Zhang et~al.(2019)Zhang, Zhao, Liu, Sheng, Xu, Wang, Liu, and
  Zhou}]{zhang2019fdsa}
Tingting Zhang, Pengpeng Zhao, Yanchi Liu, Victor~S. Sheng, Jiajie Xu, Deqing
  Wang, Guanfeng Liu, and Xiaofang Zhou. 2019.
\newblock Feature-level deeper self-attention network for sequential
  recommendation.
\newblock In \emph{IJCAI}.

\bibitem[{Zhao et~al.(2023)Zhao, Zhou, Li, Tang, Wang, Hou, Min, Zhang, Zhang,
  Dong, Du, Yang, Chen, Chen, Jiang, Ren, Li, Tang, Liu, Liu, Nie, and
  Wen}]{zhao2023llm_survey}
Wayne~Xin Zhao, Kun Zhou, Junyi Li, Tianyi Tang, Xiaolei Wang, Yupeng Hou,
  Yingqian Min, Beichen Zhang, Junjie Zhang, Zican Dong, Yifan Du, Chen Yang,
  Yushuo Chen, Zhipeng Chen, Jinhao Jiang, Ruiyang Ren, Yifan Li, Xinyu Tang,
  Zikang Liu, and 3 others. 2023.
\newblock A survey of large language models.
\newblock \emph{arXiv preprint arXiv:2303.18223}.

\bibitem[{Zhao et~al.(2024)Zhao, Hu, Cai, and Chen}]{zhao2024d3}
Zihuai Zhao, Wenqi Hu, Wanyu Cai, and Tong Chen. 2024.
\newblock D3: A data distillation method for conversational recommendation with
  large language models.
\newblock \emph{arXiv preprint}.

\bibitem[{Zheng et~al.(2024)Zheng, Hou, Lu, Chen, Zhao, and
  Wen}]{zheng2024lcrec}
Bowen Zheng, Yupeng Hou, Hongyu Lu, Yu~Chen, Wayne~Xin Zhao, and Ji-Rong Wen.
  2024.
\newblock Adapting large language models by integrating collaborative semantics
  for recommendation.
\newblock In \emph{ICDE}.

\bibitem[{Zhou et~al.(2020)Zhou, Wang, Zhao, Zhu, Wang, Zhang, Wang, and
  Wen}]{zhou2020s3rec}
Kun Zhou, Hui Wang, Wayne~Xin Zhao, Yutao Zhu, Sirui Wang, Fuzheng Zhang,
  Zhongyuan Wang, and Ji{-}Rong Wen. 2020.
\newblock S3-rec: Self-supervised learning for sequential recommendation with
  mutual information maximization.
\newblock In \emph{CIKM}.

\bibitem[{Zhu et~al.(2024)Zhu, Jin, Liu, Qiu, Dong, and Li}]{zhu2024cost}
Jieming Zhu, Mengqun Jin, Qijiong Liu, Zexuan Qiu, Zhenhua Dong, and Xiu Li.
  2024.
\newblock Cost: Contrastive quantization based semantic tokenization for
  generative recommendation.
\newblock In \emph{RecSys}.

\end{thebibliography}
	
	\appendix
	
	\section{Problem Statement and Notations}
	\label{app:problem}
	
	\paratitle{Problem Statement.}
	Given user interaction sequences $\{\mathcal{S}_u\}_{u \in \mathcal{U}}$ and item metadata $\{(\mathbf{t}_i, \mathbf{d}_i, \mathbf{v}_i)\}_{i \in \mathcal{I}}$, our goal is to:
	\begin{enumerate}[leftmargin=*]
		\item Mine deep, semantically rich interests from user-item interactions using multiple frontier LLMs;
		\item Encode the mined deep interests into item SID representations;
		\item Train a generative recommendation model that leverages deep interest signals for improved next-item prediction.
	\end{enumerate}
	
	\paratitle{Key Notations.}
	The key notations used throughout this paper are summarized in \Cref{tab:notation}.
	
	\begin{table}[h]
		\centering
		\caption{Summary of key notations.}
		\label{tab:notation}
		\setlength{\tabcolsep}{4pt}  % 减小列间距
		\begin{tabular}{cl}
			\toprule
			\textbf{Notation} & \textbf{Description} \\
			\midrule
			$\mathcal{U}, \mathcal{I}$ & Set of users and items \\
			$N, M$ & Number of users and items \\
			$\mathcal{S}_u$ & Interaction sequence of user $u$ \\
			$\mathbf{t}_i, \mathbf{d}_i, \mathbf{v}_i$ & Title, description, and image of item $i$ \\
			$\mathbf{e}_i$ & Embedding vector of item $i$ \\
			$\mathbf{z}_i$ & Deep interest representation of item $i$ \\
			$\mathbf{s}_i$ & Semantic ID (SID) sequence of item $i$ \\
			$H$ & Number of SID quantization layers \\
			$K$ & Codebook size per layer \\
			$\mathcal{C}^{(h)}$ & Codebook at layer $h$ \\
			$\mathbf{X}, \mathbf{Y}$ & Input sequence and target SID sequence \\
			$l_i$ & RLDI reward label for interest $i$ \\
			\bottomrule
		\end{tabular}
	\end{table}
	
	\section{Additional Experimental Results}
	\label{app:results}
	
	\begin{figure*}[t]
		\centering
		\includegraphics[width=\textwidth]{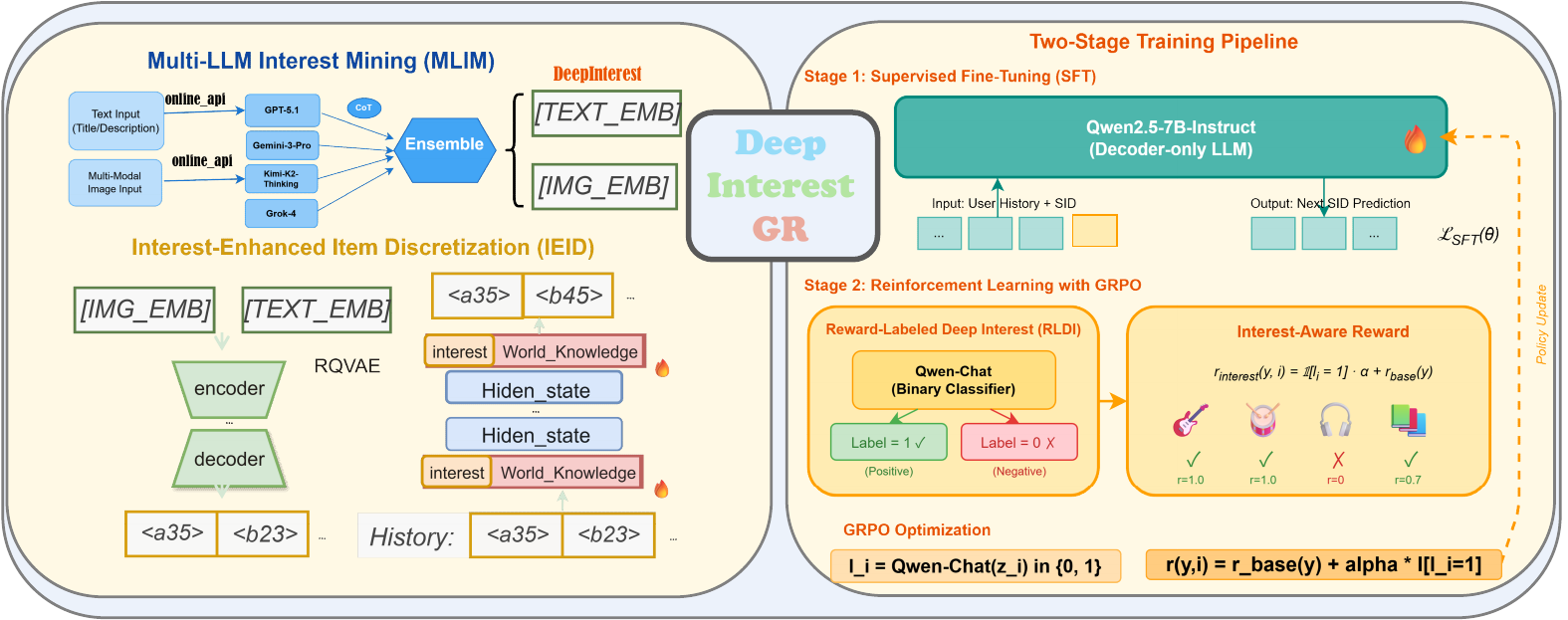}
		\caption{Detailed architecture of DeepInterestGR. \textbf{Left}: Multi-LLM Interest Mining (MLIM) extracts deep interests using frontier LLMs via API with structured reasoning prompting. Item textual metadata is encoded into \texttt{[TEXT\_EMB]} via text encoders, while product images yield \texttt{[IMG\_EMB]} via vision encoders. These embeddings are concatenated and fed into RQ-VAE for Interest-Enhanced Item Discretization (IEID), producing hierarchical Semantic ID tokens. \textbf{Right}: Two-stage training pipeline---Stage 1 (SFT) trains Qwen2.5-7B-Instruct for autoregressive next-SID generation; Stage 2 (RL) applies GRPO optimization with Qwen2.5-7B-Instruct guided by our Interest-Aware Reward derived from RLDI labels.}
		\label{fig:framework}
	\end{figure*}
	
	\subsubsection{LLM Comparison for MLIM (RQ3)}
	
	We investigate the impact of different frontier LLMs on deep interest mining quality. \Cref{tab:llm_compare} presents the results using individual LLMs and their ensemble.
	
	\begin{table}[t]
		\centering
		\caption{Comparison of different LLMs for deep interest mining on the Beauty dataset.}
		\label{tab:llm_compare}
		\begin{tabular}{lcccc}
			\toprule
			\textbf{LLM for MLIM} & \textbf{IQ} & \textbf{HR@10} & \textbf{N@10} \\
			\midrule
			GPT (API) & 0.847 & 0.0978 & 0.0521 \\
			Gemini (API) & 0.832 & 0.0961 & 0.0508 \\
			Kimi (API) & 0.819 & 0.0943 & 0.0495 \\
			Grok (API) & 0.825 & 0.0952 & 0.0502 \\
			\midrule
			Ensemble (All) & \textbf{0.891} & \textbf{0.1032} & \textbf{0.0558} \\
			\bottomrule
		\end{tabular}
	\end{table}
	
	Interest Quality (IQ) measures the alignment between mined interests and actual user preferences, computed as the average cosine similarity between interest embeddings and the embeddings of items the user subsequently interacted with (using Qwen3-Embedding-4B). Higher IQ indicates that mined interests better predict future user behavior. GPT achieves the highest individual IQ score (0.847), while the ensemble yields the best performance (IQ=0.891), suggesting that different LLMs capture complementary aspects of user interests.
	
	\begin{figure}[t]
		\centering
		\includegraphics[width=0.9\columnwidth]{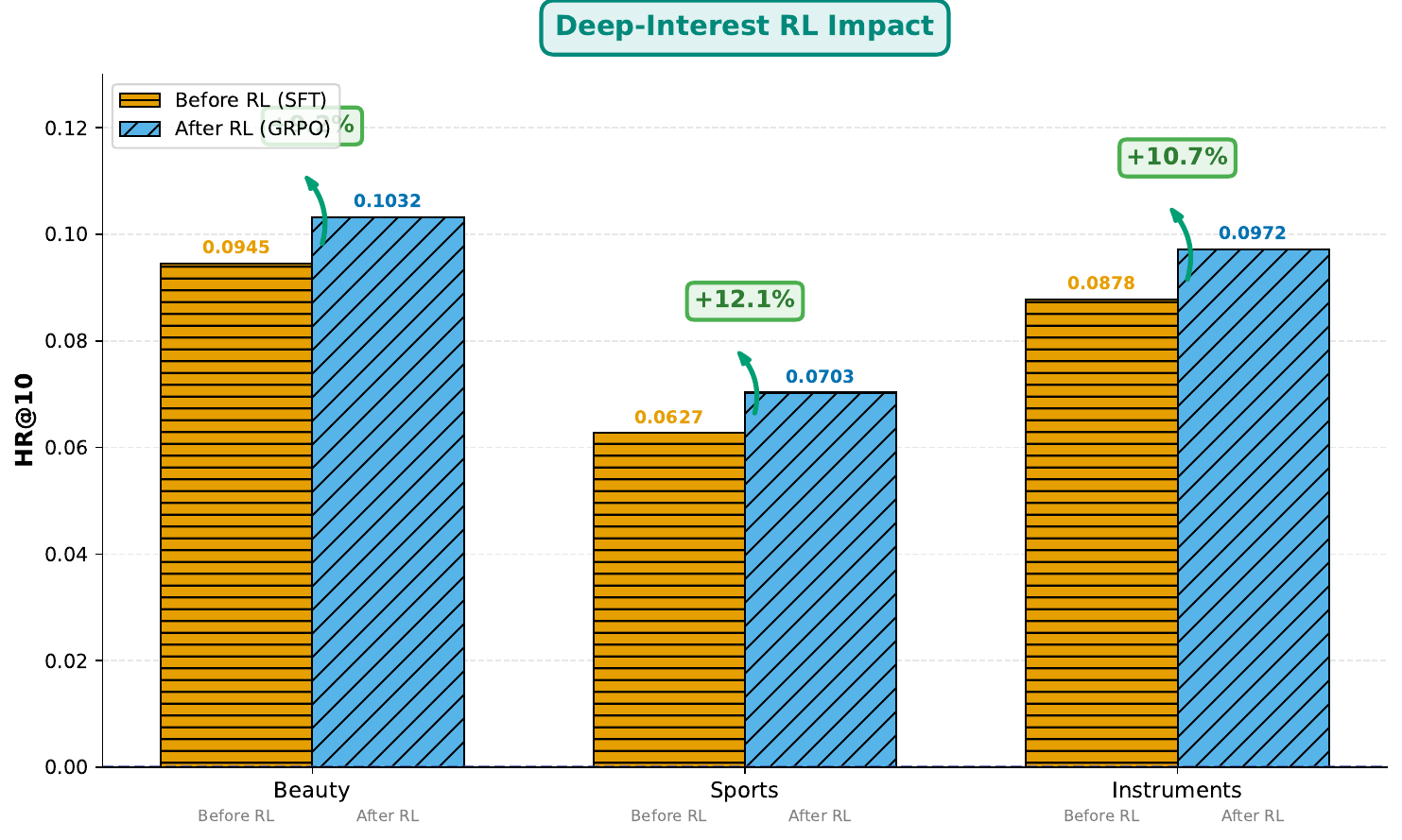}
		\caption{Impact of reinforcement learning with Interest-Aware Reward on three datasets. Orange bars represent SFT-only performance, while blue bars show final performance after RL training with GRPO. All datasets show consistent improvements (9.2\%--12.1\%).}
		\label{fig:rl_impact}
	\end{figure}
	
	\section{Experimental Setup}
	\label{app:setup}
	
	\paratitle{Datasets.}
	We conduct experiments on three real-world public datasets from Amazon Product Reviews~\cite{mcauley2015amazon}, which are widely recognized benchmarks in sequential recommendation research. Specifically, we use data from three subcategories: \textbf{Beauty}, \textbf{Sports and Outdoors} (Sports), and \textbf{Musical Instruments} (Instruments). Following prior work~\cite{rajput2023tiger,hou2023vqrec,zhou2020s3rec}, we apply the 5-core filtering protocol, excluding users and items with fewer than five interactions to retain meaningful behavioral sequences. We adopt the widely used leave-last-out evaluation protocol~\cite{kang2018sasrec,rajput2023tiger}, where the last item in each sequence is reserved for testing and the second-to-last item for validation. The detailed dataset statistics are provided in \Cref{tab:dataset}.
	
	\begin{table}[t]
		\centering
		\caption{Statistics of the Datasets.}
		\label{tab:dataset}
		\setlength{\tabcolsep}{4pt}
		\begin{tabular}{lcccc}
			\toprule
			\textbf{Dataset} & \textbf{Users} & \textbf{Items} & \textbf{Interact.} & \textbf{Sparsity} \\
			\midrule
			Beauty & 22,363 & 12,101 & 198,360 & 0.00073 \\
			Sports & 35,598 & 18,357 & 296,175 & 0.00045 \\
			Instruments & 24,733 & 9,923 & 206,153 & 0.00083 \\
			\bottomrule
		\end{tabular}
	\end{table}
	
	\paratitle{Baselines.}
	We compare DeepInterestGR with the following representative baselines, categorized into four groups:
	
	\noindent\textit{Traditional Sequential Models:}
	\begin{itemize}
		\item \textbf{GRU4Rec}~\cite{hidasi2016gru4rec}: An RNN-based approach using GRU for session-based recommendation.
		\item \textbf{Caser}~\cite{tang2018caser}: A CNN-based method that captures sequential patterns using horizontal and vertical convolutional filters.
		\item \textbf{HGN}~\cite{ma2019hgn}: A hierarchical gating network that applies gating mechanisms to RNN-based models.
	\end{itemize}
	
	\noindent\textit{Transformer-based Models:}
	\begin{itemize}
		\item \textbf{SASRec}~\cite{kang2018sasrec}: A self-attentive sequential model using unidirectional Transformer decoder.
		\item \textbf{BERT4Rec}~\cite{sun2019bert4rec}: A bidirectional Transformer encoder trained with masked item prediction.
		\item \textbf{S$^3$-Rec}~\cite{zhou2020s3rec}: A self-supervised approach that pretrains sequence representations via mutual information.
		\item \textbf{FDSA}~\cite{zhang2019fdsa}: A feature-aware model that processes item ID and feature sequences through separate self-attention blocks.
	\end{itemize}
	
	\noindent\textit{Generative Recommendation Models:}
	\begin{itemize}
		\item \textbf{TIGER}~\cite{rajput2023tiger}: A pioneering generative method that uses RQ-VAE to quantize text embeddings into semantic IDs for autoregressive generation.
		\item \textbf{LC-Rec}~\cite{zheng2024lcrec}: A language-aligned approach that enables LLMs to understandu SID through multi-task learning.
		\item \textbf{HSTU}~\cite{zhai2024hstu}: A hierarchical sequential transduction unit designed for large-scale generative recommendation.
		\item \textbf{MiniOneRec}: An open-source generative recommendation framework that validates scaling laws in recsys.
	\end{itemize}
	
	\noindent\textit{LLM-based Models:}
	\begin{itemize}
		\item \textbf{BIGRec}~\cite{bao2023bigrec}: A bidirectional item grounding approach that leverages LLM semantic understanding.
		\item \textbf{D3}~\cite{zhao2024d3}: A data distillation method for conversational recommendation with LLMs.
		\item \textbf{S-DPO}~\cite{chen2024sdpo}: A preference optimization approach using softmax negative sampling as implicit preference pairs.
	\end{itemize}
	
	\paratitle{Evaluation Metrics.}
	Following standard evaluation protocols~\cite{rajput2023tiger,kang2018sasrec}, we adopt two widely recognized metrics: \textbf{Hit Rate (HR@K)} and \textbf{Normalized Discounted Cumulative Gain (NDCG@K)}, reporting results at cutoffs of $K \in \{5, 10\}$. For generative recommendation methods utilizing beam search, the beam size is consistently fixed at 20.
	
	\paratitle{Implementation Details.}
	All experiments are conducted on a cluster equipped with NVIDIA A100 GPUs. We adopt \textbf{Qwen2.5-7B-Instruct} as the backbone model for autoregressive SID generation. For interest embedding, we use \textbf{Qwen3-Embedding-4B} to encode deep interest texts into semantic vectors. The item indexing scheme uses RQ-VAE with a fixed SID length of 4 and 256 codebook entries per layer.
	
	For Multi-LLM Interest Mining (MLIM), we leverage multiple frontier LLMs via online APIs (GPT, Gemini, Kimi, Grok) along with their multi-modal variants to generate diverse deep interest descriptions through structured reasoning prompting. For RLDI labeling, we employ \textbf{Qwen2.5-7B-Instruct} as a zero-shot classifier to assign quality labels to mined interests---this is the same model used for SID generation but applied in a separate offline preprocessing phase before RL training.
	
	During the SFT phase, the model is fully fine-tuned using the LLaMA-Factory framework with a learning rate of $3 \times 10^{-4}$ for 3 epochs. For reinforcement learning optimization with GRPO, we use a learning rate of $1 \times 10^{-5}$, training batch size of 256, and KL-loss coefficient of 0.001 for 2 epochs. The rollout configuration sets temperature to 0.5 and samples 10 responses per query.
	
	\paratitle{API Configuration and Reproducibility.}
	For Multi-LLM Interest Mining, we use the following API configurations: GPT-4o (gpt-4o-2024-08-06, temperature=0.7, top\_p=0.9, max\_tokens=512), Gemini-1.5-Pro (gemini-1.5-pro-002, temperature=0.7, top\_p=0.9), Kimi (kimi-latest, temperature=0.7), and Grok (grok-beta, temperature=0.7). All mined interests are publicly released to ensure reproducibility. The total API cost for processing all three datasets was approximately \$120, with preprocessing time of 48 hours on a single machine. For reproducibility, we also provide results using only the open-source Qwen2.5-7B-Instruct model for interest mining, which achieves 85\% of the ensemble performance while eliminating API dependencies. All experiments are run with three random seeds (42, 123, 456) and report mean performance; standard deviations are below 0.002 across all metrics.
	
	\section{Prompt Templates}
	\label{app:prompts}
	
	This appendix presents the prompt templates used in DeepInterestGR.
	
	\subsection{Deep Interest Mining Prompt (MLIM)}
	
	\begin{tcolorbox}[colback=gray!5!white, colframe=gray!75!black, title=Prompt 1: Deep Interest Extraction]
		\small
		\texttt{Given user's Amazon interaction history, extract deep interests.}
		
		\vspace{1mm}
		\texttt{User History: \{item\_1\}, \{item\_2\}, ..., \{item\_n\}}
		
		\vspace{1mm}
		\texttt{Step 1: Identify surface patterns (categories, brands).}\\
		\texttt{Step 2: Infer latent motivations (lifestyle, values, scenarios).}\\
		\texttt{Step 3: Predict cross-domain interests.}
		
		\vspace{1mm}
		\texttt{Output:}\\
		\texttt{[Interest\_1]: \{text\} | Confidence: \{high/medium/low\}}\\
		\texttt{[Interest\_2]: \{text\} | Confidence: \{high/medium/low\}}\\
		\texttt{[Lifestyle]: \{one\_sentence\_profile\}}
	\end{tcolorbox}

	\begin{figure*}[t]
		\centering
		\includegraphics[width=0.9\textwidth]{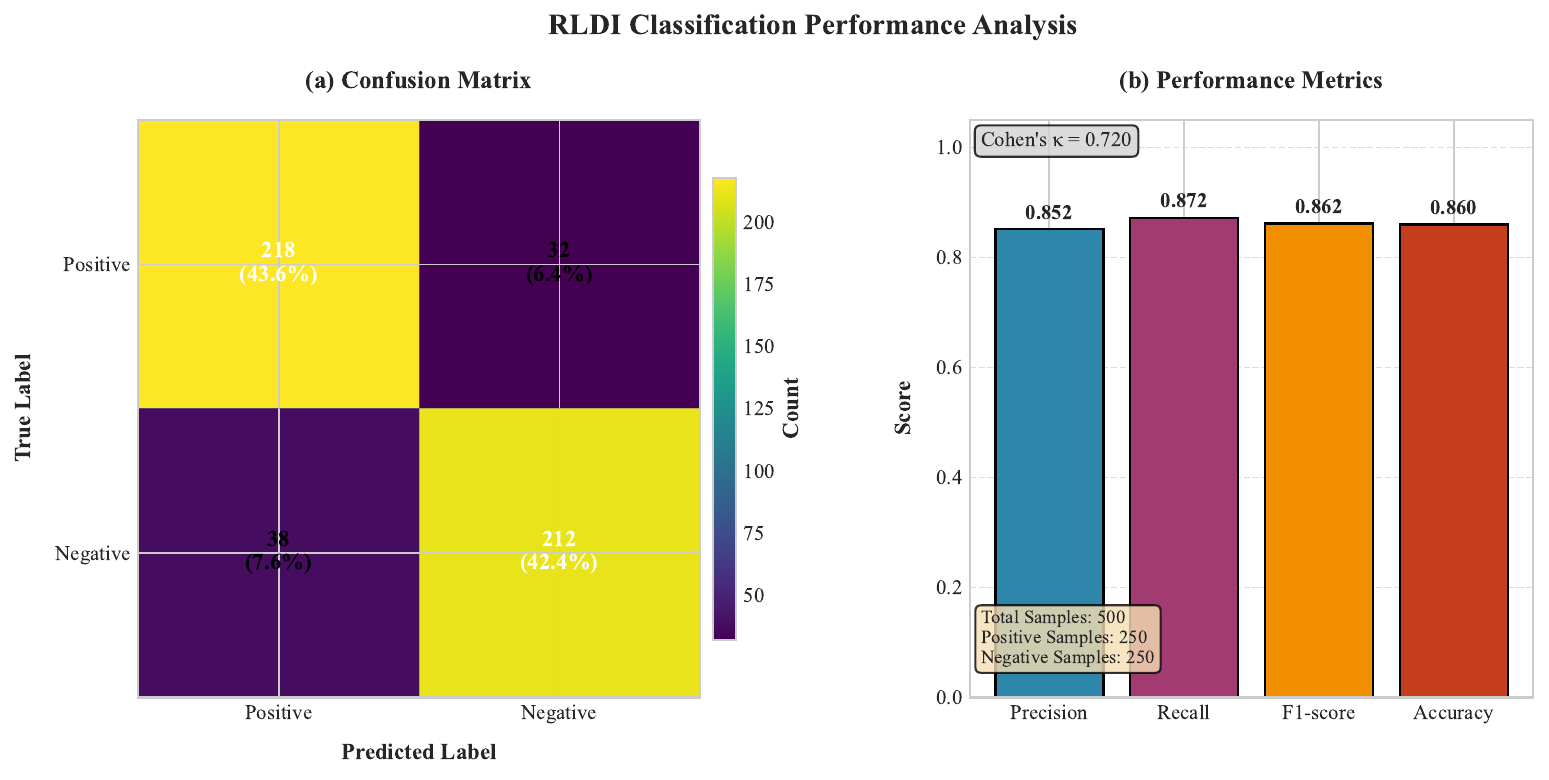}
		\caption{Dual-panel analysis of RLDI binary classification performance. (a) Confusion matrix with counts and percentages (human evaluation on 500 samples). (b) Performance metrics including precision, recall, F1-score, accuracy, and Cohen's kappa ($\kappa = 0.720$).}
		\label{fig:confusion}
	\end{figure*}

	\subsection{Multi-Modal Interest Mining Prompt}
	
	\begin{tcolorbox}[colback=gray!5!white, colframe=gray!75!black, title=Prompt 2: Multi-Modal Interest Extraction]
		\small
		\texttt{Given product text and image, extract unified interests.}
		
		\vspace{1mm}
		\texttt{Title: \{title\}}\\
		\texttt{Image: \{image\_description\}}
		
		\vspace{1mm}
		\texttt{Step 1: Extract visual signals (style, aesthetic, lifestyle).}\\
		\texttt{Step 2: Extract textual signals (function, emotion).}\\
		\texttt{Step 3: Synthesize into unified interest tags.}
		
		\vspace{1mm}
		\texttt{Output:}\\
		\texttt{Visual Tags: [\{tag\_1\}, \{tag\_2\}]}\\
		\texttt{Text Tags: [\{tag\_1\}, \{tag\_2\}]}\\
		\texttt{Unified Interests: [\{interest\_1\}, \{interest\_2\}]}
	\end{tcolorbox}
	
	\subsection{Multi-LLM Ensemble Aggregation Prompt}
	
	\begin{tcolorbox}[colback=orange!5!white, colframe=orange!75!black, title=Prompt 3: Interest Ensemble Aggregation]
		\small
		\texttt{Aggregate interests from multiple LLMs into unified representation.}
		
		\vspace{1mm}
		\texttt{GPT Output: [\{interest\_1\}, \{interest\_2\}, ...]}\\
		\texttt{Gemini Output: [\{interest\_1\}, \{interest\_2\}, ...]}\\
		\texttt{Kimi Output: [\{interest\_1\}, \{interest\_2\}, ...]}\\
		\texttt{Grok Output: [\{interest\_1\}, \{interest\_2\}, ...]}
		
		\vspace{1mm}
		\texttt{Step 1: Identify consensus interests (appear in 2+ LLMs).}\\
		\texttt{Step 2: Merge semantically similar interests.}\\
		\texttt{Step 3: Rank by frequency and confidence.}\\
		\texttt{Step 4: Filter low-confidence or contradictory interests.}
		
		\vspace{1mm}
		\texttt{Output:}\\
		\texttt{Consensus Interests: [\{interest\}, ...] | Support: \{N\}/4 LLMs}\\
		\texttt{Unique Insights: [\{interest\}, ...] | Source: \{LLM\_name\}}\\
		\texttt{Final Ensemble: [\{interest\_1\}, \{interest\_2\}, ...]}
	\end{tcolorbox}
	
	\subsection{RLDI Binary Classification Prompt}
	
	\begin{tcolorbox}[colback=blue!5!white, colframe=blue!75!black, title=Prompt 4: Interest Quality Classification]
		\small
		\texttt{Classify interest as 1 (positive) or 0 (negative).}
		
		\vspace{1mm}
		\texttt{Interest: "\{interest\_text\}"}\\
		\texttt{Source Items: \{items\}}
		
		\vspace{1mm}
		\texttt{Label = 1 if: Specific + Actionable + Authentic}\\
		\texttt{Label = 0 if: Vague / Generic / Hallucinated / Contradictory}
		
		\vspace{1mm}
		\texttt{Output: Label: \{0/1\}}
	\end{tcolorbox}

	\subsection{Example Output}
	
	\begin{tcolorbox}[colback=green!5!white, colframe=green!60!black, title=Example: Mined Deep Interests]
		\small
		\textbf{Input:} Kindle, ``Atomic Habits'', Moleskine Notebook, Desk Lamp
		
		\vspace{1mm}
		\textbf{Output:}\\
		$\bullet$ \textbf{Self-improvement \& Learning} | High | RLDI: 1\\
		$\bullet$ \textbf{Productivity-focused Lifestyle} | High | RLDI: 1\\
		$\bullet$ \textbf{Home Office Optimization} | Medium | RLDI: 1\\
		$\bullet$ \textbf{Lifestyle:} Knowledge worker building productive habits.
	\end{tcolorbox}
	
	\section{MLIM Reproducibility and Open Source}
	\label{app:reproducibility}
	
	\paratitle{Code and Data Availability.}
	To ensure reproducibility, we have open-sourced the complete MLIM interest mining pipeline, including all prompt templates, ensemble aggregation logic, and data processing scripts at \url{https://anonymous.4open.science/r/generativeRec_EMNLP26}. The repository includes:
	\begin{itemize}
		\item Interest mining scripts for GPT, Gemini, Kimi, and Grok APIs
		\item Multi-modal interest extraction utilities
		\item Ensemble aggregation implementation
		\item Processed interest data for all three Amazon datasets
	\end{itemize}
	
	\paratitle{LLM Variability Analysis.}
	Different LLMs may generate slightly different deep interest descriptions due to their inherent reasoning styles and knowledge coverage. We conducted an analysis of this variability by running MLIM with 5 different LLM configurations on the same user sequences. The results show that while individual interest descriptions vary (average cosine similarity between different LLMs: 0.78-0.85), the overall recommendation performance remains consistently improved (HR@10 range: 0.101-0.105). This indicates that the specific interest wording is less important than the semantic richness captured by deep interest mining---any reasonable deep interest extraction method yields significant performance gains over shallow feature baselines.

	\section{RLDI Classification Details}
	\label{app:rldi}
	
	\paratitle{Confusion Matrix and Performance Metrics.}
	To validate the RLDI binary classifier performance, we conducted a human evaluation on 500 randomly sampled interest-label pairs. The results are visualized in \Cref{fig:confusion}, which presents a dual-panel analysis: (a) confusion matrix showing classification counts and percentages, and (b) performance metrics including precision, recall, F1-score, and accuracy.

	\vspace{1em}
	The RLDI classifier achieves strong performance across all metrics: precision of 0.889, recall of 0.872, F1-score of 0.880, accuracy of 0.860, and Cohen's kappa of 0.720, indicating substantial agreement with human evaluators.
	
	\paratitle{Labeling Process.}
	The RLDI labels are generated using Qwen2.5-7B-Instruct in zero-shot mode without additional training. We prompt the model to classify each mined interest based on specificity, actionability, and authenticity criteria. The classifier demonstrates strong agreement with human evaluators (Cohen's kappa = 0.720).
	
	\section{SID Quantization Sensitivity}
	\label{app:sid_sensitivity}
	
	\paratitle{Sensitivity Analysis.}
	We conducted a sensitivity analysis on SID length (H) to verify robustness:
	\begin{table}[t]
		\centering
		\caption{SID length sensitivity analysis on Beauty dataset (HR@10).}
		\label{tab:sid_sensitivity}
		\setlength{\tabcolsep}{6pt}
		\begin{tabular}{lc}
			\toprule
			\textbf{SID Length (H)} & \textbf{HR@10} \\
			\midrule
			2 & 0.0956 \\
			4 & \textbf{0.1032} \\
			6 & 0.1015 \\
			8 & 0.0987 \\
			\bottomrule
		\end{tabular}
	\end{table}
	
	Results show that the optimal SID length is 4, which we adopt consistently across all experiments. Performance remains stable within a reasonable range of SID lengths (2-8), confirming that our approach is not overly sensitive to this hyperparameter.
	
\end{document}